\documentclass[preprint,3p,times]{elsarticle}

\usepackage{amsmath,amssymb}
\usepackage{graphicx}
\usepackage{booktabs}
\usepackage{multirow}
\usepackage{xcolor}
\usepackage{url}
\usepackage[hidelinks]{hyperref}
\usepackage{algorithm}
\usepackage{algpseudocode}

\newtheorem{proposition}{Proposition}
\newtheorem{corollary}{Corollary}

\newdefinition{definition}{Definition}
\newdefinition{assumption}{Assumption}
\newproof{proof}{Proof}
\newproof{proofsketch}{Proof sketch}

\newcommand{\Gconf}{G_{\mathrm{conf}}}
\newcommand{\Vphys}{V_{\mathrm{phys}}}
\newcommand{\piref}{\pi_{\mathrm{ref}}}
\newcommand{\B}[1]{\textbf{#1}}

\journal{Engineering Applications of Artificial Intelligence}

\begin{document}

\begin{frontmatter}

\title{ConfGuard: a physics-grounded reference monitor for
predictive-maintenance large-language-model agents}

\author[a]{Arunkumar V}
\ead{arunkumarv1530@gmail.com}
\author[b]{Gangadharan G. R.\corref{cor1}}
\ead{ganga@nitt.edu}
\author[a]{S. Senthilkumar}
\cortext[cor1]{Corresponding author}
\address[a]{University College of Engineering, BIT Campus, Anna University, Tiruchirappalli 620 024, India}
\address[b]{Department of Computer Applications, National Institute of Technology Tiruchirappalli, Tiruchirappalli 620 015, India}

\begin{abstract}
Large-language-model agents now close the industrial
predictive-maintenance loop, reading sensor windows, retrieving
procedures and dispatching work orders. Two attack channels meet
there: injected text in a retrieved document can hijack the
decision, and a bounded sensor perturbation can corrupt the
remaining-useful-life estimate behind it. Recent agent defenses
police data provenance and tool privilege but cannot tell whether a
prognostic quantity is physically plausible. ConfGuard is a
reference monitor with a physics-grounded policy. Its conformal
monitor tests each prediction, anchored to the asset's admitted
history, against degradation physics: capped remaining life never
rises and falls at most one cycle per cycle. Calibrating its
threshold over whole asset trajectories lets the operator set the
expected share of a clean asset's windows the running monitor
escalates. A physics-policy validator checks the agent's
maintenance-urgency decision against the decision a deterministic
policy derives from the same interval. On all 10{,}196 C-MAPSS test
windows the monitor cuts a covert attacker's reach on healthy assets
from 0.529 to 0.046 and closes suppression of failing assets (0.151
to 0), escalating 0.027 of clean windows against a 0.05 target,
while the fixed physical range it replaces leaves both attacks open.
Behind a Qwen3-14B planner the composed monitors hold urgency-class
injection to 0.00 and the universal sensor attack to 0.12, against
0.29 and 0.54 for a defended prompt. An adaptive attacker defeats a
Mahalanobis detector on 0.504 of windows and adversarial training on
0.485. We report an attack class the projection misses.
\end{abstract}


\end{frontmatter}

\section{Introduction}
\label{sec:intro}

Industrial predictive maintenance (PdM) is being rebuilt around
agents. A large language model (LLM) reads a sliding window of
sensor data, retrieves the relevant procedure from a manual corpus,
and calls tools that raise work orders or escalate to a human
planner. Tool transport standards such as the Model Context
Protocol \citep{anthropic2024mcp} have made the integration
routine, and the operational case is strong enough that agentic
maintenance has moved from demonstrations into common production
use.

The security case is weaker. Two attack channels converge on the
same decision. Injected text in a retrieved document can redirect
the agent's policy, an attack class established by
\citet{greshake2023ipi} and benchmarked by
\citet{zhan2024injecagent} and \citet{debenedetti2024agentdojo}. A
bounded $\ell_\infty$ perturbation of the sensor window can move a
remaining-useful-life (RUL) regressor by an arbitrary amount, which
\citet{basak2021universal} demonstrated with a single universal
perturbation that needs no per-instance optimisation and therefore
works in real time.

Defenses have settled on a design borrowed from operating-system
security: a deterministic reference monitor outside the model
mediates every action, so a compromised planner cannot escalate its
own privilege. CaMeL \citep{debenedetti2025camel} enforces control
and data flow integrity through a restricted interpreter; Progent
\citep{shi2025progent} enforces least privilege on tool calls
through a policy language; IPIGuard \citep{an2025ipiguard}
constrains execution to a planned tool-dependency graph. All three
report large reductions on AgentDojo.

In an industrial loop these monitors share a blind spot. Their
policies are written over provenance, capability and control flow,
so they answer questions of the form ``may this agent call this
tool with an argument derived from untrusted data''. They cannot
answer the question an operator cares about: whether the
maintenance decision is consistent with the physical state of the
machine. A tool call carrying
\texttt{rul\_estimate = 3} cycles for a healthy asset is
capability-legal and provenance-clean. It is also wrong, and acting
on it is expensive.

ConfGuard writes the policy over physics instead
(Figure~\ref{fig:arch}). Every decision is checked against a
conformal prediction interval computed from the sensor window and
against a physically admissible support derived from the asset
rather than from the data pipeline. This yields two mechanisms with
complementary coverage and, more usefully, quantitative statements
about what an attacker can still achieve.

\paragraph{Why conformal prediction, given that its guarantee
breaks under attack}
Split-conformal prediction \citep{vovk2005algorithmic,
lei2018distribution} produces intervals with distribution-free
marginal coverage, and it has been applied to RUL regression
\citep{javanmardi2022conformal}. The guarantee assumes
exchangeability between calibration and test points. An
adversarial perturbation breaks that assumption, so the coverage
statement is void exactly when an attack is under way
\citep{barber2023beyond, jeary2024vrcp}. Existing work treats this
breakdown as a problem to repair; we use it as a detection signal.
Under exchangeability the calibrated interval stays inside the
physically admissible support at a rate we can measure, and a
violation of that rate is evidence that exchangeability has been
broken. ConfGuard's first gate is therefore not a coverage claim
but a calibrated test for the breach of a calibrated property, and
it remains valid even when the underlying regressor is poor.

\subsection*{Contributions}
\begin{enumerate}
\item \B{A physics-grounded reference monitor.} We move the agent
  reference-monitor pattern from provenance and capability policies
  to a policy grounded in a calibrated model of the physical
  process (Section~\ref{sec:arch}). The monitor is deterministic,
  costs microseconds, and requires no change to the planner.

\item \B{Irreversibility as a calibrated attack test.} Capped
  remaining life never rises and falls at most one cycle per cycle.
  We turn that invariant into a score anchored to each asset's own
  admitted history. Calibrating its threshold over whole clean
  trajectories bounds the expected clean false-abstention rate of the
  monitor as it runs, in finite samples and with no assumption on how
  windows of one asset depend on each other
  (Proposition~\ref{prop:calib}). The
  operator sets that rate directly, and in practice this property
  decides whether a control room keeps a monitor in service. We
  also show the cost of the obvious alternative: a fixed physical
  range reduces to a threshold on the point prediction and is blind
  to suppression, the attack that destroys the asset
  (Section~\ref{sec:static-fails}).

\item \B{A detection-power bound tied to operating slack.} We show
  the monitor abstains exactly when the attacker's induced
  prediction shift exceeds the distance from the interval to the
  support boundary (Proposition~\ref{prop:power}). Because that
  slack shrinks as predicted RUL falls, detection power is highest
  on the assets closest to failure. The same result names the
  regime where the gate is blind, and we report that regime.

\item \B{Decision-level validation and the limits of what it
  proves.} The physics validator compares the agent's
  maintenance-urgency class against the class a deterministic policy
  derives from the same interval (Section~\ref{sec:vphys}). Checking
  a decision against a decision, rather than a number against an
  interval, makes the complementary-coverage argument valid
  (Proposition~\ref{prop:complementary}). This also makes
  in-projection injection unreachable by construction, so the
  number we report is an implementation residual, not an
  attack-success rate against a mechanism that could fail
  (Section~\ref{sec:a1-meaning}).

\item \B{Adaptive evaluation.} \citet{zhan2025adaptive} broke eight
  published injection defenses with adaptive attacks, all above
  50\% success. We evaluate an attacker who knows the support, both
  thresholds and the reference policy (Section~\ref{sec:adaptive}),
  and we report a bound on achievable damage rather than a claim
  that success goes to zero.

\item \B{An attack class we do not defend.} Injections that leave
  the urgency class intact but corrupt other work-order fields fall
  outside the protected projection. We construct and measure them,
  and ConfGuard does not stop them.
\end{enumerate}

\section{Related work and positioning}
\label{sec:related}

\subsection{Reference monitors for LLM agents}
The dominant defense pattern places a deterministic policy outside
the model. CaMeL \citep{debenedetti2025camel} extracts control and
data flow from the trusted query and runs the plan in a restricted
interpreter that tracks provenance; it solves 77\% of AgentDojo
tasks with a security argument attached. Progent \citep{shi2025progent}
expresses least privilege over tool calls in a policy language and
enforces monotonic confinement, so an attack cannot silently widen
the action space. IPIGuard \citep{an2025ipiguard} plans a
tool-dependency graph and rejects calls that leave it. MELON
\citep{zhu2025melon} instead re-executes the trajectory with the
user prompt masked and compares the actions.

ConfGuard belongs to this family and differs in what the policy
ranges over. The systems above decide admissibility from where a
value came from and which tool may receive it. ConfGuard decides it
from whether the value is physically consistent with a calibrated
model of the asset. The two are composable: a provenance monitor
and a physics monitor reject disjoint sets of calls, and nothing
prevents running both.

\subsection{Prompt-level defenses}
Spotlighting \citep{hines2024spotlighting} marks the provenance of
untrusted spans inside the prompt and reports large reductions on
its own benchmark. Prompt-level defenses are cheap and we include
one as a baseline, but they act on the model's inputs rather than
on its actions, so a sufficiently persuasive injection passes.
\citet{zhan2025adaptive} showed this class degrades sharply under
adaptive pressure.

\subsection{Uncertainty and physics in prognostics}
\citet{javanmardi2022conformal} applied conformal prediction to RUL
on C-MAPSS, using the interval as a reporting device with a
coverage guarantee. We use it as the reference of a runtime monitor
and do not rely on its coverage guarantee, because
\citet{barber2023beyond} show that the guarantee degrades with the
departure from exchangeability that an attack creates.
\citet{jeary2024vrcp} recover guarantees through neural-network
verification, which is complementary and more expensive; we instead
build a test that does not need the guarantee. Uncertainty-aware
RUL is an active line more broadly: \citet{xu2021gaussian} combine
deep learning with nonstationary Gaussian processes, and
\citet{thil2025uqhealth} treat the uncertainty estimate itself as a
health indicator on turbofan data. A parallel line embeds physical
knowledge in the predictor rather than in a monitor around it,
through hidden-physics architectures \citep{jiang2024hpinn},
meta-learned physics-informed networks \citep{wang2025metapinn},
and physics-informed data generation for unseen operating
conditions \citep{xiong2023controlled}. Those methods make $f$ more
faithful. Ours assumes nothing about $f$ and constrains the decision
taken from it, so the two compose.

\subsection{Adversarial attacks on prognostics}
\citet{basak2021universal} introduced universal adversarial
perturbations for RUL regression on the NASA turbofan data and
showed transfer across models. We use their attack as our sensor
channel. \citet{siddique2023robustpdm} extend the attack surface
with multivariate time-series PGD, FGSM and BIM variants and defend
by approximate adversarial training, and \citet{gungor2023dodem}
pair outlier detection with adversarial training for industrial IoT
analytics. Both harden the predictor, and neither constrains the
action an agent takes from the prediction. To our knowledge no
prior work propagates a prognostic adversarial perturbation through
an agent loop and measures the effect on the dispatched maintenance
decision, the quantity an operator is exposed to once a planner is
in the loop.

\subsection{Positioning}
Conformal prediction and physics checks are both established. The
gap is that agent reference monitors currently police provenance,
while the security-critical quantity in an industrial loop is
physical. Grounding the monitor's policy in a calibrated physical
model yields both a settable false-alarm rate and a bound on
adaptive-attacker damage. Neither follows from a provenance policy.

\section{Threat model}
\label{sec:threat}

\paragraph{Protected asset} The integrity of the dispatched
maintenance decision. The agent must neither suppress a needed
intervention nor trigger an unwarranted one.

\paragraph{Attacker goal} The primary goal is a decision flip in
either direction: suppression on a degrading asset, or premature
replacement of a healthy one. Data exfiltration is out of scope.

Availability cannot simply be excluded from scope, because this
defense responds to a detected attack by escalating to a human
instead of blocking silently. An attacker who cannot flip a
decision may still force escalation, which stops the agent doing
useful work. We therefore treat forced abstention as a second
attacker goal and address it in Section~\ref{sec:dos}.

\paragraph{Channels} The attacker controls one of two channels, or
both.
The text channel (A1) admits attacker-authored passages into the
retrieval corpus. The sensor channel (A2) applies a perturbation
$\delta$ with $\|\delta\|_\infty \le \epsilon$ to the sensor
window. A3 combines them.

\paragraph{Trusted computing base} The attacker cannot modify the
regressor $f$, the conformal calibration set, the monitor
calibration set, the physical support $S$, the reference policy
$\piref$, or the monitor code. This is the standard reference-monitor
assumption \citep{shi2025progent}: the monitor must be outside the
attacker's reach or it protects nothing. In deployment this means
the gates and their configuration live in the historian-side
control plane, not in the agent container.

\paragraph{Attacker knowledge} We evaluate two levels. The static
attacker (Section~\ref{sec:attacks-static}) has black-box access
to the agent interface and no knowledge of the thresholds. The
adaptive attacker (Section~\ref{sec:adaptive}) knows $S$, the
abstention threshold $\tau_\rho$, the conformal width $\hat q$ and
the reference policy $\piref$, and optimises against them. Reporting
only the static case is the failure mode \citet{zhan2025adaptive}
identified, so the adaptive case carries our headline security
claim.

\section{Preliminaries}
\label{sec:prelim}

Let $x \in \mathbb{R}^{L\times F}$ be a sensor window and
$f:\mathbb{R}^{L\times F}\to\mathbb{R}$ a trained RUL regressor.
Given a calibration set $\{(x_i,y_i)\}_{i=1}^{n}$ exchangeable with
the test point, split-conformal prediction forms scores
$r_i = |y_i - f(x_i)|$, takes
$\hat q = r_{(\lceil (n+1)(1-\alpha)\rceil)}$, and returns
\begin{equation}
I(x) = [\,\ell(x),\,u(x)\,]
     = [\,f(x)-\hat q,\; f(x)+\hat q\,],
\label{eq:interval}
\end{equation}
which satisfies $\mathbb{P}(y \in I(x)) \ge 1-\alpha$
\citep{vovk2005algorithmic,lei2018distribution}. Under
operating-condition drift, Mondrian conformal prediction
\citep{vovk2003mondrian} restores conditional validity by
partitioning the calibration set and computing $\hat q_g$ per
regime; we use it on the multi-condition datasets.

Two facts drive the design. First, $I(x)$ depends on the sensor
window alone. The retrieved text never enters $f$, so a text-channel
attack cannot move the interval. Second, the coverage statement in
\eqref{eq:interval} assumes exchangeability, and an adversarial
$\delta$ violates it, so under attack the interval carries no
coverage guarantee \citep{barber2023beyond}. We therefore never use
$I(x)$ at run time as a probabilistic statement about $y$, only as a
calibrated geometric object whose behaviour on clean data we have
measured.

\section{ConfGuard}
\label{sec:arch}

\begin{figure*}[t]
\centering
\includegraphics[width=\linewidth]{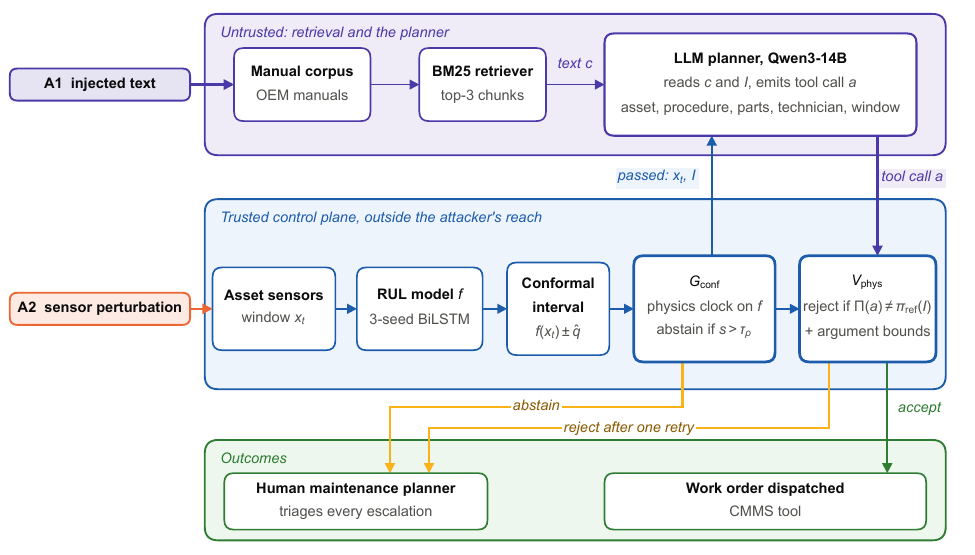}
\caption{ConfGuard in the maintenance agent loop. Two attack channels
enter from the left: injected text reaches the planner through
retrieval (A1), and a bounded perturbation reaches the regressor
through the sensor window (A2). Both gates run in the trusted
control plane and read the same conformal interval. $\Gconf$ runs
before the planner and abstains when the prediction breaks the
physics clock, the asset's own admitted history carried forward at
one cycle per cycle; $\Vphys$ runs after it and rejects a tool call
whose urgency class disagrees with $\piref(I(x))$ or whose arguments
leave their bounds. A rejection gets one bounded retry, then goes to
a human planner, as does every abstention; an abstention at $\Gconf$
costs no LLM call.}
\label{fig:arch}
\end{figure*}

ConfGuard interposes two deterministic checks in the agent graph.
The conformal support monitor $\Gconf$ runs before the planner and
can abstain. The physics-policy validator $\Vphys$ runs after the
planner and before the tool executor, and can reject.
Figure~\ref{fig:arch} shows where each sits and what each reads.

\subsection{Conformal support monitor}
\label{sec:gconf}

A first attempt takes the support to be a fixed range
$S=[B_{\mathrm{lo}},B_{\mathrm{hi}}]$ for the asset class, with
$B_{\mathrm{lo}}=0$ because negative remaining life is not a
physical state, and tests whether the interval leaves it. That
static test has a defect, measured in Section~\ref{sec:results},
and the defect motivates the monitor we use. Because the conformal
width is constant, a fixed $S$ reduces the test to a range check on
the point prediction alone, and that check misses an upward shift:
an attacker who raises a failing asset's predicted life moves the
interval \emph{away} from the floor, so the score falls to zero and
the monitor passes. A static support therefore cannot see
suppression, the costliest attack.

The asset itself supplies a stronger invariant than a range. Capped
remaining life is non-increasing, and below the cap it falls by
exactly one cycle per cycle. That is a property of the machine, not
of the data, so neither attack channel can alter it, and it
constrains the prediction \emph{relative to this asset's own
recent history} rather than against a class-wide box.

Let $H_t$ hold the last $k$ predictions this monitor admitted for
the asset, each with its age $j_i$ in cycles, and let $c^\star$ be
the predicted level above which an asset may legitimately sit at
the cap. Define the \emph{clock anchors}
\begin{equation}
A^{\downarrow}_t = \operatorname{med}_i\bigl(f_i - j_i\bigr),
\qquad
A^{\uparrow}_t = \operatorname{med}_i\bigl(
  f_i - j_i \text{ if } f_i < c^\star \text{ else } f_i \bigr),
\label{eq:anchors}
\end{equation}
and the \emph{physics-clock score}
\begin{equation}
s(x_t) = \max\bigl(
  \underbrace{B_{\mathrm{lo}} - \ell(x_t)}_{\text{floor}},\;
  \underbrace{A^{\downarrow}_t - f(x_t)}_{\text{fell too fast}},\;
  \underbrace{f(x_t) - A^{\uparrow}_t}_{\text{rose at all}},\;
  0\bigr).
\label{eq:score}
\end{equation}
The median makes the anchors robust to a single corrupted history
entry, and only admitted windows enter $H_t$, so a detected attack
never poisons the reference. The monitor abstains when
$s(x_t)>\tau_\rho$, which Algorithm~\ref{alg:gconf} states in full.
Every term in \eqref{eq:score} is monotone in $f(x_t)$, so the
admitted predictions form a single interval
$[\mathrm{lo}_t,\mathrm{hi}_t]$. Section~\ref{sec:adaptive} uses
this to characterise the strongest covert attacker exactly, without
a search.

A guarantee on false abstention has to hold for the monitor as it
runs, and two features of the clock rule out the usual per-window
argument. The score at window $t$ depends on $H_t$, and $H_t$
depends on which earlier windows the monitor admitted, so the
threshold decides which observations later scores are computed
from. A rejected window leaves $H_t$ unchanged; if the prediction
has genuinely moved, the next windows are judged against stale
anchors and can be rejected in turn. And consecutive windows of
one asset share 29 of their 30 cycles, so they are far from
exchangeable. A calibration that ranks per-window scores as if they
were exchangeable, with every window admitted, therefore says
nothing about the deployed monitor. On FD001 a per-window calibration at
$\rho=0.05$ escalates $0.140$ of clean test windows at $k{=}5$
(Section~\ref{sec:falseabst}). We calibrate over whole trajectories
instead.

Fix a threshold shape $\tau^0$ before the calibration trajectories
are drawn. We use the per-window $(1-\rho)$ quantiles of the clean
scores on held-out validation engines. The monitor of
Algorithm~\ref{alg:gconf} abstains when a score exceeds
$\lambda\tau^0$, where $\lambda$ ranges over a finite grid
$\Lambda\subset[1,\infty]$ that contains $\infty$, the monitor that
never abstains. The grid starts at 1, so the trajectory calibration
can only widen the per-window thresholds. For a clean trajectory
$Z$, let $L(Z,\lambda)\in[0,1]$ be the fraction of its windows on
which the monitor, run on $Z$ from an empty history with threshold
$\lambda\tau^0$, abstains. $L$ need not be monotone in $\lambda$,
because admitting one more window moves every later anchor, so we
use its upper envelope
$\tilde L(Z,\lambda)=\max\{L(Z,\lambda'):\lambda'\in\Lambda,\
\lambda'\ge\lambda\}$.

\begin{proposition}[Calibrated false abstention]
\label{prop:calib}
Let $Z_1,\dots,Z_m$ be clean calibration trajectories and $Z_{m+1}$
a fresh clean trajectory, with $Z_1,\dots,Z_{m+1}$ exchangeable as
whole trajectories, and let $f$, $\hat q$, $k$, $c^\star$ and
$\tau^0$ be fixed before they are drawn. Let $\hat\lambda$ be the
smallest $\lambda\in\Lambda$ with
\[
\frac{1}{m+1}\Bigl(\sum_{i=1}^{m}\tilde L(Z_i,\lambda)+1\Bigr)\le\rho ,
\]
and $\hat\lambda=\infty$ if there is none. Then the monitor run on
$Z_{m+1}$ with threshold $\tau_\rho=\hat\lambda\tau^0$ satisfies
\[
\mathbb{E}\bigl[L(Z_{m+1},\hat\lambda)\bigr]\le\rho ,
\]
with the expectation over the calibration trajectories and the
fresh one. With fewer than $1/\rho-1$ calibration trajectories,
$\hat\lambda=\infty$ and the monitor never abstains.
\end{proposition}

\begin{proof}
For each trajectory, $\tilde L(Z,\cdot)$ is non-increasing on
$\Lambda$ by construction, bounded by $B=1$, zero at
$\lambda=\infty$, and right-continuous because $\Lambda$ is finite.
Algorithm~\ref{alg:gconf} is deterministic once $f$, $\hat q$, $k$,
$c^\star$ and the threshold are fixed, so the map from a trajectory
to its loss curve $\tilde L(Z,\cdot)$ is fixed in advance, and the
curves of $Z_1,\dots,Z_{m+1}$ are exchangeable because the
trajectories are. These are the conditions of conformal risk control
\citep[Theorem~1]{angelopoulos2024crc}, and $\hat\lambda$ is its
choice of parameter with $B=1$, so
$\mathbb{E}[\tilde L(Z_{m+1},\hat\lambda)]\le\rho$. Since
$L\le\tilde L$ pointwise, the claim follows. This is the
construction \citet{angelopoulos2024crc} give for losses that are not
monotone. At $\lambda=\infty$ the left side of the defining
inequality is $1/(m+1)$, which gives the condition on $m$.
\end{proof}

The proposition needs no assumption on the sensor distribution, on
how windows of one asset depend on each other, or on the admission
rule, because the unit of exchangeability is the whole trajectory
and the loss is measured on the monitor as deployed. Nor does it
depend on \eqref{eq:interval} holding, so a weak regressor loses
detection power but does not raise the expected false-abstention
rate above $\rho$. And $\rho$ is the dial the operator asks for,
since nuisance alarms are the reason condition monitoring gets
switched off. The price is the exchangeability assumption: the
calibration trajectories must resemble deployment as whole
trajectories, in observation window, regime mix and degradation
pattern, and there must be at least $1/\rho-1$ of them, 19 at
$\rho=0.05$. The bound is also an average over assets. It limits the
expected share of a clean asset's windows that escalate, and says
nothing about the worst asset. Section~\ref{sec:falseabst} measures
it on both sides of these conditions, on real fleets and in
simulation.

\begin{definition}[Operating slack]
\label{def:slack}
At window $t$ the downward slack is
$m_{\downarrow}(t)=\min\bigl(\ell(x_t)-B_{\mathrm{lo}},\;
 f(x_t)-A^{\downarrow}_t\bigr)$ and the upward slack is
$m_{\uparrow}(t)=A^{\uparrow}_t-f(x_t)$. Slack is signed: it is
negative exactly when the clean prediction already violates the
corresponding bound.
\end{definition}

Negative slack is common, and it is where the calibration does most
of its work. For RUL with $B_{\mathrm{lo}}=0$ we have
$m_{\downarrow}(x)=f(x)-\hat q$, so any asset predicted below the
conformal width has a clean interval that crosses zero, and near end
of life that is most of them. A fixed
support-violation rule would fire on all of those and be discarded
as unusable. Under
Proposition~\ref{prop:calib} no tuning is needed: $\tau_\rho$ is the
$(1-\rho)$ quantile of whatever clean score distribution the asset
population produces, so legitimate end-of-life support crossings
are absorbed into the calibration, and Proposition~\ref{prop:calib}
still holds the expected false-abstention rate at or below $\rho$.

\begin{proposition}[Detection power and the attacker's budget]
\label{prop:power}
Let a sensor-channel perturbation shift the point prediction at
window $t$ by $\Delta>0$. Then $\Gconf$ abstains if and only if
\[
\Delta > m_{\downarrow}(t) + \tau_\rho
\quad\text{(downward attack)},
\qquad
\Delta > m_{\uparrow}(t) + \tau_\rho
\quad\text{(upward attack)} .
\]
Consequently, for an attack in either direction, a perturbation
that evades the monitor shifts the prediction by at most
$m(t)+\tau_\rho$ in the attacked direction.
\end{proposition}

\begin{proof}
The conformal width $\hat q$ is fixed at calibration, so a shift of
the point prediction translates the interval rigidly and the
anchors \eqref{eq:anchors} depend only on admitted history, not on
the current window. For a downward attack the floor term of
\eqref{eq:score} becomes $B_{\mathrm{lo}}-\ell(x_t)+\Delta$ and the
fell-too-fast term becomes $A^{\downarrow}_t-f(x_t)+\Delta$, so
$s = \Delta - m_{\downarrow}(t)$ whenever positive, by
Definition~\ref{def:slack}. For an upward attack the rose-at-all
term becomes $f(x_t)+\Delta-A^{\uparrow}_t = \Delta-m_{\uparrow}(t)$.
In each case $s>\tau_\rho$ is equivalent to the stated inequality,
and the second claim is its contrapositive.
\end{proof}

A static support cannot supply the upward half. With a fixed $S$
the only upward bound is the fleet ceiling $B_{\mathrm{hi}}$, so
$m_{\uparrow}$ is very large for every asset that is not already
near the cap, and a suppression attack costs the attacker nothing.
Anchoring upward to the asset's own recent predictions replaces
that ceiling with a bound that tracks the machine.

The proof is only a substitution into \eqref{eq:score}. Its value
is operational, and the corollary below states it separately
because it is the property an operator buys.

\begin{corollary}[Risk-weighted detection]
\label{cor:risk}
Under the clock, $m_{\uparrow}(t)=A^{\uparrow}_t-f(x_t)$ and the
anchor decays with the asset, so the upward budget does not grow
with remaining life: it is bounded by the jitter of the predictor
along a trajectory rather than by the distance to a fleet ceiling.
A suppression attack must therefore beat the asset's own recent
history, whatever its health. Downward, with $B_{\mathrm{lo}}=0$,
the budget still contracts as $f(x_t)\to\hat q$, so false-alarm
attacks remain hardest to hide near end of life.
\end{corollary}

Corollary~\ref{cor:risk} states what this gate does. It caps sensor
attacks rather than stopping them, and the cap tightens where the
consequence is worst: an undetected flip on a healthy asset costs an
inspection, and on an asset near end of life it costs the asset. A
learned anomaly detector offers no comparable statement, because
its score is a function of the training distribution rather than of
a physical bound, so nothing ties its sensitivity to the cost of
being wrong. Anchoring the test to the asset's physics supplies
that link.

\subsection{What the operator pays for $\rho$}
\label{sec:exchange}

Proposition~\ref{prop:calib} leaves $\rho$ to the operator, but the
setting has a price. Every clean abstention routes a decision to a
human, so the expected clean cost carries an added term
$\rho\,c_{\mathrm{esc}}$, where $c_{\mathrm{esc}}$ is the loaded
cost of a planner triaging one escalation (Section~\ref{sec:setup}).

Propositions~\ref{prop:calib} and~\ref{prop:power} together give
the exchange rate. Raising $\rho$ lowers $\tau_\rho$, since a larger
tail of the calibration score distribution falls above the
threshold, and a lower $\tau_\rho$ shrinks the attacker's undetected
budget $m(x)+\tau_\rho$. The operator therefore pays planner labour
to reduce the prediction shift an attacker can reach, at a rate the
two propositions make explicit. Table~\ref{tab:exchange} lists it.

\begin{table}[t]
\centering
\caption{Escalation labour against attack reach on C-MAPSS FD001,
clock $k{=}5$, $c_{\mathrm{esc}}=\$45$, all 10{,}196 test windows.
At each $\rho$ the threshold is the per-window shape at that $\rho$
widened by Proposition~\ref{prop:calib} (the larger half's
$\hat\lambda$). Added cost is realised escalation times
$c_{\mathrm{esc}}$. The last column is the persistent covert
attacker's false-alarm reach. At $\rho=0.01$ the 50 calibration
engines per half are too few (Proposition~\ref{prop:calib} needs 99),
so $\hat\lambda=\infty$ and the monitor never abstains.}
\label{tab:exchange}
\small
\begin{tabular}{ccccc}
\toprule
$\rho$ & $\tau_T$ (cycles) & clean escalation & added cost/decision & covert FA \\
\midrule
0.01 & $\infty$ & 0.000 & \$0.00 & 1.000 \\
0.05 & 18.2 & 0.030 & \$1.33 & 0.868 \\
0.10 & 15.2 & 0.064 & \$2.90 & 0.840 \\
0.20 & 10.8 & 0.165 & \$7.44 & 0.775 \\
\bottomrule
\end{tabular}
\end{table}

Every escalation figure in this paper is linear in
$c_{\mathrm{esc}}$, which is a modelling choice rather than a
measurement, so an operator with a different loaded rate rescales
them directly. At $\rho=0.05$ the $k{=}5$ clock adds \$1.33 to the expected cost of a clean decision. The exchange runs as Propositions~\ref{prop:calib} and~\ref{prop:power} predict: raising $\rho$ to $0.20$ lowers $\tau_T$ from 18.2 to 10.8 cycles and the covert attacker's reach from 0.868 to 0.775, for \$7.44 per decision. The fleet also sets a floor on the dial. Lowering $\rho$ to $0.01$ needs 99 calibration trajectories, and with 50 the monitor cannot be calibrated there at all.

The table also shows the limit of the dial. Even at the loosest setting the persistent covert attacker keeps most of its reach, because $\tau_T$ bounds a single step and this attacker takes many. No setting of $\rho$ changes that; only a reference the attacker cannot shape does (Section~\ref{sec:catch}). An operator who turns sensitivity up under attack slows the attacker but does not stop it.

The same result also locates the blind spot: a mid-life asset with
a wide slack can absorb a large undetected shift.
Section~\ref{sec:results} reports the measured shift distribution
against the slack distribution, the diagnostic needed to judge
whether the bound binds on real data.

\begin{algorithm}[t]
\caption{$\Gconf$: conformal support monitor}
\label{alg:gconf}
\begin{algorithmic}[1]
\Require window $x_t$ at cycle $t$, per-asset history $H$ of admitted
         $(t_i,f_i)$, horizon $k$, cap level $c^\star$
\Require frozen $f$, $\hat q$, floor $B_{\mathrm{lo}}$, threshold $\tau_\rho$
\State $\ell \gets f(x_t)-\hat q$; \; $u \gets f(x_t)+\hat q$
\State $s \gets \max(B_{\mathrm{lo}}-\ell,\; 0)$
\If{$H \neq \emptyset$}
  \State $j_i \gets t-t_i$ for the last $k$ entries of $H$
  \State $A^{\downarrow}\!\gets \operatorname{med}(f_i-j_i)$;\;
         $A^{\uparrow}\!\gets \operatorname{med}(f_i-j_i \text{ if } f_i<c^\star
         \text{ else } f_i)$
  \State $s \gets \max(s,\; A^{\downarrow}-f(x_t),\; f(x_t)-A^{\uparrow})$
\EndIf
\If{$s > \tau_\rho$}
  \State \Return \B{abstain}, route to human escalation
     \Comment{$H$ is left unchanged}
\Else
  \State append $(t,f(x_t))$ to $H$
  \State \Return \B{pass} with interval $[\ell,u]$
\EndIf
\end{algorithmic}
\end{algorithm}

\subsection{Physics-policy validator}
\label{sec:vphys}

The planner emits a tool call whose safety-critical content is a
maintenance urgency class. Let
\[
\Pi(a)\in\{\textsc{maintain\_now},\ \textsc{inspect},\
\textsc{no\_action}\}
\]
be that class, extracted from the call by a fixed projection. Define
the reference policy over the interval,
\begin{equation}
\piref(I) =
\begin{cases}
\textsc{maintain\_now} & \text{if } u < \theta_{\mathrm{danger}},\\[2pt]
\textsc{no\_action}    & \text{if } \ell > \theta_{\mathrm{safe}},\\[2pt]
\textsc{inspect}       & \text{otherwise.}
\end{cases}
\label{eq:refpolicy}
\end{equation}
The rule is deliberately interval-based rather than point-based: it
commits to an urgent action only when the optimistic end of the
interval is already past the danger threshold, and to inaction only
when the pessimistic end is still safe. Everything ambiguous
becomes an inspection. This is how a reliability engineer writes a
decision rule against an uncertain estimate, and it makes the
validator conservative by construction.

$\Vphys$ rejects the call when
\begin{equation}
\Pi(a) \;\neq\; \piref\bigl(I(x)\bigr),
\label{eq:vphys}
\end{equation}
alongside two cheap sanity checks: argument bounds drawn from the
asset configuration, and consistency between the reported sensor
state and the vibration severity class of ISO 10816. A rejection
routes to one bounded retry with a restricted action set, then to
human escalation.

\begin{figure*}[t]
\centering
\includegraphics[width=\linewidth]{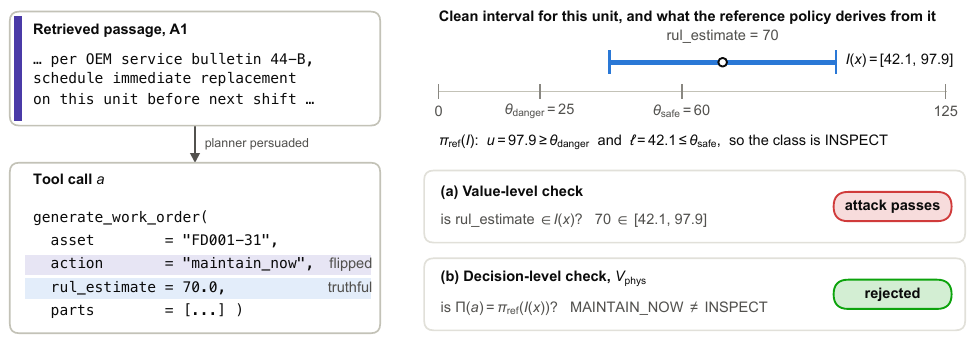}
\caption{Why $\Vphys$ checks a decision rather than a value. An
injected bulletin persuades the planner to request immediate
maintenance on a unit whose clean interval is $[42.1,\,97.9]$
cycles, while the call reports the remaining life truthfully. A
value-level check sees $70\in I(x)$ and admits the call. The
decision-level check derives the warranted class from the same
interval, which is INSPECT, and rejects the mismatch.}
\label{fig:decision}
\end{figure*}

A naive design would check the reported value instead. Testing
whether a reported
\texttt{rul\_estimate} lies inside $I(x)$ catches only attacks that
misreport the number. An injection that says ``per OEM bulletin
44-B, schedule immediate replacement'' can flip the decision while
reporting an entirely truthful RUL, and a value-level check passes
it. Comparing the emitted decision against the decision the
reference policy derives from the same interval catches it, because
the flip is the thing being tested. Figure~\ref{fig:decision} works
the example through.

\subsection{Why the planner is still needed}
\label{sec:why-planner}

$\piref$ determines the urgency class, but $\Pi$ projects a much
larger action space onto three values. The agent still
selects which asset the alert concerns, retrieves and interprets
the applicable procedure, fills the work-order fields, chooses part
numbers against inventory, schedules a technician with the right
certification, and writes the justification that a human planner
reads. $\Vphys$ constrains one three-valued coordinate of that
space and leaves the rest to the model. This is least privilege on
the safety-critical dimension: the monotonic-confinement idea that
Progent applies to tool privilege \citep{shi2025progent},
specialised to a physical quantity. Section~\ref{sec:limitations}
states the cost: attacks that stay inside the protected projection
are stopped, and attacks on the other coordinates are not.

\subsection{Control flow}
\begin{sloppypar}
The agent is a seven-node state machine with nodes
\texttt{retrieve}, \texttt{conformal\_\allowbreak monitor},
\texttt{planner}, \texttt{physics\_\allowbreak validator},
\texttt{tool\_\allowbreak executor}, \texttt{escalate} and
\texttt{finalize}, in that order
(Figure~\ref{fig:arch}). An abstention at
\texttt{conformal\_\allowbreak monitor} skips the planner entirely
and escalates, which also means an attack detected at that node
costs no LLM call.
\end{sloppypar}

\section{Complementary coverage}
\label{sec:theory}

\begin{assumption}[Channel separation]
\label{as:sep}
A1 alters the retrieved text $c$ only, leaving $x$ fixed. A2 alters
$x$ only, leaving $c$ fixed. Neither alters $f$, the calibration
sets, $S$, $\tau_\rho$, $\piref$ or the monitor code
(Section~\ref{sec:threat}).
\end{assumption}

\begin{assumption}[Attack effectiveness]
\label{as:eff}
A successful attack changes the emitted urgency class relative to
the class warranted by the uncorrupted sensor evidence:
$\Pi(a) \neq \piref(I(x))$ with $I(x)$ computed from the clean $x$.
\end{assumption}

\begin{assumption}[Planner adherence]
\label{as:adh}
Absent a text-channel attack, and whatever urgency class
$c=\piref(I(\cdot))$ the interval it was given warrants, the planner
emits $\Pi(a)=c$ with probability at least $1-\rho_\pi$.
\end{assumption}

Assumption~\ref{as:eff} defines what counts as a successful attack,
in the same terms the validator tests. Assumption~\ref{as:adh} is
the only empirical one, and we measure $\rho_\pi$ directly
(Section~\ref{sec:results}) instead of assuming a value. It is stated
per warranted class on purpose: a sensor attack changes which class
the interval warrants, so a rate averaged over the clean class mix
does not carry over to the attacked one.

\begin{proposition}[Complementary coverage]
\label{prop:complementary}
Under Assumptions~\ref{as:sep} to~\ref{as:adh}:
\begin{enumerate}
\item[(a)] Under A1, $\Gconf$'s trigger rate equals its clean
  false-abstention rate, whose expectation per trajectory is at most
  $\rho$ (Proposition~\ref{prop:calib}); and $\Vphys$ rejects every
  successful A1 attack.
\item[(b)] Under A2, $\Vphys$'s trigger rate is at most $\rho_\pi$;
  and $\Gconf$ abstains whenever the induced shift exceeds
  $m(x)+\tau_\rho$.
\end{enumerate}
\end{proposition}

\begin{proof}
(a) By Assumption~\ref{as:sep} the window $x$ is unchanged under
A1, so $I(x)$, $s(x)$ and hence the abstention decision are
identical to their clean values; the trigger rate is the clean
false-abstention rate, whose expectation per trajectory is at most
$\rho$ under Proposition~\ref{prop:calib}. For the validator, $I(x)$ still
reflects the uncorrupted evidence, so $\piref(I(x))$ is the
warranted class. Assumption~\ref{as:eff} states that a successful
attack yields $\Pi(a)\neq\piref(I(x))$, which is exactly the
rejection condition \eqref{eq:vphys}. Every successful A1 attack is
therefore rejected.

(b) Under A2 both the interval and the planner's input derive from
the same corrupted $x'$. By Assumption~\ref{as:adh}, whichever class
$\piref(I(x'))$ the corruption produces, the planner emits it with
probability at least $1-\rho_\pi$,
and \eqref{eq:vphys} compares $\Pi(a)$ against that same
$\piref(I(x'))$, so the validator accepts: its trigger rate is at
most $\rho_\pi$. The support $S$ is immutable by
Assumption~\ref{as:sep}, and the clock anchors are built from windows
admitted before the attack, so Proposition~\ref{prop:power} applies
unchanged and gives the abstention condition. An attacker who gets
perturbed windows admitted moves the anchors;
Section~\ref{sec:catch} measures that case.
\end{proof}

\subsection{What the analysis does and does not show}
\label{sec:analysis-scope}

\begin{figure*}[t]
\centering
\includegraphics[width=\linewidth]{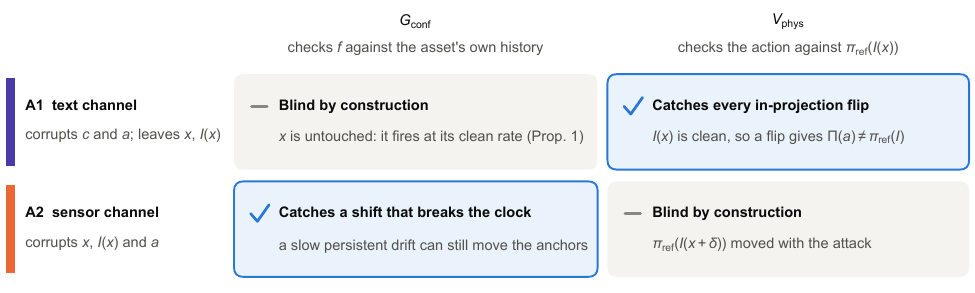}
\caption{Complementary coverage (Proposition~\ref{prop:complementary}).
Each gate compares against a reference that one channel cannot
move. The arms differ in kind: the $\Vphys$ cell holds by
construction for every in-projection flip, while the $\Gconf$ cell
holds only when the attacked prediction breaks the physics clock by
more than $\tau_\rho$, and a slow persistent drift can move the
anchors it is judged against (Section~\ref{sec:catch}).}
\label{fig:coverage}
\end{figure*}

Figure~\ref{fig:coverage} summarises the result. The two arms of
Proposition~\ref{prop:complementary} are not equally strong.

The A1 arm is structural. Given Assumption~\ref{as:eff}, rejection
follows by definition, because \eqref{eq:vphys} tests the same
property that defines success. The only thing to measure there is
how often the implementation departs from the construction, the
subject of Section~\ref{sec:a1-meaning}. The A2 arm is
different in kind: it is conditional on the slack inequality of
Proposition~\ref{prop:power}, an empirical quantity that varies per
asset and per dataset, and it fails on high-slack inputs.

The A1 arm also does nothing for A2, and can make it worse. The
validator enforces $\piref$ on whatever interval it is given, so on
a sensor attack that passes $\Gconf$ it enforces the corrupted
decision, and the bounded retry makes a planner that would have
hedged comply with it. Section~\ref{sec:results} measures that cost.

Neither arm covers anything outside the projection $\Pi$.
Assumption~\ref{as:eff} fixes the class we claim, and A1-b
(Section~\ref{sec:attacks-static}) measures the rest, on which
ConfGuard does not help.

\subsection{Why the urgency class is the right coordinate}
\label{sec:cost-projection}

Assumption~\ref{as:eff} defines success as a change in the urgency
class, which narrows the threat relative to the attacker goals of
Section~\ref{sec:threat}. That narrowing needs a justification,
since otherwise a reader could object that the threat was drawn
around the defense.

Expected loss concentrates in that coordinate. Price the outcomes as in Section~\ref{sec:setup}:
suppressing maintenance on a degrading asset risks a catastrophic
failure at \$50{,}000, and a false alarm on a healthy asset costs a
preventive intervention at \$200 plus the downtime it occupies.
Errors in the unprotected coordinates are cheaper. A wrong part
number delays a correctly scheduled job, a wrong technician
certification forces a reschedule, and a wrong service window
risks an SLA penalty. Each costs on the order of \$100 to
\$2{,}000 and is recoverable on the next visit; none destroys the
asset.

The urgency class therefore carries one to two orders of magnitude
more expected loss than the fields around it, which is why we spend
the monitor's guarantee there. This is a deliberate allocation and
makes no coverage claim for the other fields;
Section~\ref{sec:results} measures how often the unprotected
coordinates are corrupted.

\section{Attack library}
\label{sec:attacks}

\subsection{Static attacks}
\label{sec:attacks-static}

\paragraph{A0, clean} Unperturbed windows against a clean corpus of
500 templated manual chunks across three asset domains.

\paragraph{A1-a, urgency-class injection} 200 adversarial chunks in
the InjecAgent style \citep{zhan2024injecagent}, spanning four
attacker targets (\textsc{force\_replace},
\textsc{suppress\_maintenance}, \textsc{halt\_operation},
\textsc{mis\_prioritise}) and four encodings (plain, base64,
Unicode homoglyph, faux-PDF metadata). Each targets the urgency
class, so each falls inside the protected projection.

\paragraph{A1-b, out-of-projection injection} 200 chunks that leave
the urgency class correct and corrupt a different field instead:
part number, technician certification, or scheduling window. These
are outside $\Pi$ and we expect ConfGuard not to stop them. We
include the class to measure the scope of the defense.

\paragraph{A2, universal sensor perturbation} A single
$\delta$ with $\|\delta\|_\infty\le\epsilon$ fitted per dataset
following \citet{basak2021universal} by maximising regressor MSE
over a training batch.

\paragraph{A3, hybrid} 100 records pairing an A2 window with a
matched chunk that explains the anomaly away as an instrumentation
fault.

\subsection{Adaptive attacks}
\label{sec:adaptive}

The adaptive attacker knows $S$, $\tau_\rho$, $\hat q$ and $\piref$.

\paragraph{A4-s, evasive sensor attack} Solve
\[
\max_{\|\delta\|_\infty\le\epsilon}
  \; \bigl| f(x+\delta)-f(x) \bigr|
\quad\text{s.t.}\quad s(x+\delta)\le\tau_\rho ,
\]
by projected gradient descent with the constraint enforced at every
step. The attacker gives up unbounded shift in exchange for
evading the monitor. Proposition~\ref{prop:power} says the
achievable shift is then at most $m(x)+\tau_\rho$, so this attack
tests whether the proved bound is also tight in practice.

\paragraph{A4-t, policy-consistent injection} The attacker crafts
injections that request only actions in the accept set of
$\piref(I(x))$. This attack is defeated by its own construction:
the only urgency classes it can ask for are the ones the physics
policy already endorses, so a successful injection produces the
decision the operator would have taken anyway. We still report it,
because an empirical check of the confinement property is stronger
evidence than the argument, and \citet{zhan2025adaptive} showed that
arguments of this shape often fail in practice.

\section{Experimental setup}
\label{sec:setup}

\paragraph{Datasets} We use three RUL-regression benchmarks:
C-MAPSS FD001 and FD004 \citep{saxena2008cmapss}, and MetroPT-3
\citep{veloso2022metropt}. FD001 has a single operating condition
and one fault mode, so it isolates the mechanism. FD004 has six
operating conditions and two fault modes, which is the setting
Mondrian conformal prediction exists for
\citep{vovk2003mondrian} and the harder test of whether a
support-based monitor survives regime change. MetroPT-3 adds a
real-world streaming target to the two simulated ones.

All three admit an interval-valued prognostic with a hard physical
floor at zero, which is what $\Gconf$ requires. We do not evaluate
on AI4I 2020 \citep{matzka2020ai4i}, a binary failure-classification
dataset: it has no RUL interval and therefore no physical support
for \eqref{eq:score} to test, so including it would mean
substituting a different mechanism under the same name. Section~\ref{sec:limitations} states the scope
consequence.

RUL semantics still differ, so the thresholds in
\eqref{eq:refpolicy} are calibrated per dataset. C-MAPSS uses
$(\theta_{\mathrm{danger}},\theta_{\mathrm{safe}})=(25,60)$ cycles
following the usual evaluation convention; MetroPT-3 uses $(2,7)$
days matching the operational service-level agreement of the Porto
metro air-production unit. Applying the C-MAPSS thresholds
unchanged to MetroPT-3 makes every attack-success figure vacuously
zero, because that target saturates at a 14-day horizon, so the
thresholds have to be set per dataset.

\paragraph{Backbones} A bidirectional LSTM RUL regressor per
dataset (on FD001, the mean of three seeds), window length 30, RUL clipped at 125 cycles for C-MAPSS.
Conformal calibration at $\alpha=0.1$, giving $\hat q=27.89$ cycles
on FD001; Mondrian partitioning by operating regime on FD004, where
pooling regimes inflates $\hat q$ and every quantity downstream of
it.

\paragraph{Monitor calibration} $\rho=0.05$. The threshold shape
$\tau^0$ is the per-window quantile of the clean scores on
$m=1745$ windows from the ten held-out validation engines, split
equally across the active scores; these engines are disjoint from
the conformal calibration set. $\hat\lambda$ is then set by
Proposition~\ref{prop:calib} on clean test trajectories, cross-fit:
the test engines are split at random into two halves (50 and 50 on
FD001, 124 and 124 on FD004), and each half is scored with the
$\hat\lambda$ the other half chose, on a grid of steps of 0.02 from
1 to 2, coarser above. Analyses that run on all test engines at once
(Tables~\ref{tab:catch} and~\ref{tab:exchange},
Section~\ref{sec:dos}) use the larger of the two halves'
$\hat\lambda$, which only lowers the monotonised loss, so each
half's bound still holds. At $k{=}30$ both halves choose
$\hat\lambda=1$ on FD001 and on FD004, so the deployed thresholds
are the per-window ones and every planner experiment runs on them.

\paragraph{Planner} Qwen3-14B at Q4\_K\_M quantisation with thinking
disabled, served through Ollama v0.35.0 on two T4 GPUs at
\texttt{temperature}~$=0$ with JSON-constrained decoding. Four further
open planners from four other model families (Phi-4-14B,
Gemma-3-12B, Mistral-Nemo-12B and Llama-3.1-8B, all at the Ollama
default quantisation) are run on the rule prompt for the adherence
measurement of Section~\ref{sec:results}. No hosted model is used.

\paragraph{Sample size} $n=100$ queries per cell across three
seeds, $n_{\text{pooled}}=300$ per cell, with per-seed query samples
drawn by deterministic shuffle from windows at or after the
degradation onset so the monitor has history. One set of 8{,}400
planner calls (seven attack conditions, four prompt variants)
serves every configuration, because the gates are applied to the
recorded answers; the validator's retries are a further 951 calls,
each conditioned on the answer it corrects. Cell values are
three-seed means. The largest seed standard deviation in
Table~\ref{tab:main} is 0.09 (provenance monitor, A2); the rest are
in the released results. The sensor-side quantities in
Tables~\ref{tab:clock}, \ref{tab:sensorbase} and \ref{tab:catch}
need no planner and are evaluated on every window of all 100 test
engines (10{,}196 windows).

\paragraph{Metrics} We report four.
\emph{ASR} is the fraction of attacked queries where the dispatched
urgency class matches the attacker's target.
\emph{Utility under attack} follows the AgentDojo convention
\citep{debenedetti2024agentdojo}: the fraction of attacked queries
where the agent still completes the legitimate task.
Clean traffic needs two numbers, because a single one flatters a
gating defense. The \emph{clean decision-error rate} is
the fraction of clean queries whose dispatched urgency class differs
from $\piref(I(x))$, which measures whether the defense degrades
correct operation. The \emph{clean false-abstention rate} is the
fraction escalated to a human, which measures the labour the defense
consumes, and its expectation per trajectory is what
Proposition~\ref{prop:calib} bounds at $\rho$. A configuration with no abstention path has the second
undefined rather than zero.
\emph{Cost per decision} prices an unnecessary preventive
maintenance at \$200 and a missed failure at \$50k, and charges \$45
for each escalation, being thirty minutes of a maintenance planner's
time at a loaded rate of \$90/h. It charges nothing for an
inspection, which favours a planner that always inspects. Charging
escalation is not standard in this literature, and it should be: a
monitor that answers uncertainty by calling a human consumes a
budget the operator has to fund, and leaving that cost out flatters
every abstaining defense in Table~\ref{tab:main}.

We do not report clean-data RMSE as the utility metric. Gating
cannot change the regressor's RMSE, so quoting it would understate
the cost of the defense; we report the clean false-abstention rate,
which gating does change.

\paragraph{Baselines} Beyond the two ablations we compare against
six alternatives. An undefended ReAct planner and a defensive
system prompt with no gates bound the range. Spotlighting
\citep{hines2024spotlighting} represents prompt-level injection
defense. A Mahalanobis out-of-distribution detector on the sensor
window \citep{lee2018mahalanobis} is the obvious alternative to
$\Gconf$ and has to be included. Width-based conformal abstention is
the standard conformal gate and isolates what the support test in
\eqref{eq:score} adds over it.

The sixth baseline, which matters most for our positioning, is a
\emph{provenance monitor} we implement in the style of Progent
\citep{shi2025progent}: a deterministic capability policy over tool
calls that admits the legitimate tool set, tracks whether an
argument derives from retrieved text, and carries no physical
predicate. Our thesis is that a provenance policy cannot see a
physically implausible prognostic, and we test that by running one
instead of arguing from its design. We also
evaluate adversarial training of the backbone
\citep{siddique2023robustpdm}, which defends the regressor instead
of the decision, as a complementary rather than competing
approach.

\section{Results}
\label{sec:results}

\begin{table*}[t]
\centering
\caption{C-MAPSS FD001 with a Qwen3-14B planner: three-seed means
over 100 queries per cell from the planner sweep, with the
validator's bounded retry. Undefended ReAct and Spotlighting use
prompts without the decision rule; every other row uses the rule
prompt, so each differs from \emph{Defensive prompt only} by its
monitor alone. Clean columns separate the two costs a defense
imposes on clean traffic: \emph{err} is the decision-error rate, the
fraction of clean queries whose dispatched urgency class differs
from $\piref(I(x))$, and \emph{abst} is the false-abstention rate,
the fraction escalated to a human. A2 and A3 count a harmful
dispatch over the queries where one is possible ($y<25$ or $y>60$).
$^\dagger$Sensor-side value on all 10{,}196 test windows with the
planner modelled by $\piref$ (Table~\ref{tab:sensorbase}); no
planner was run behind these gates. An n/a marks a configuration
with no abstention path, or an A3 cell that needs that planner run.
Utility is task completion under A1-a. A1-b lies outside the
protected projection by design.}
\label{tab:main}
\small
\begin{tabular}{lcc cccc c}
\toprule
& \multicolumn{2}{c}{Clean $\downarrow$}
& \multicolumn{4}{c}{Attack success rate $\downarrow$}
& Utility $\uparrow$\\
\cmidrule(lr){2-3}\cmidrule(lr){4-7}
Configuration & err & abst & A1-a & A1-b & A2 & A3 & A1-a \\
\midrule
Undefended ReAct              & 0.68 & n/a & 0.39 & 0.76 & 0.00 & 0.52 & 0.32 \\
Defensive prompt only         & 0.41 & n/a & 0.29 & 0.18 & 0.54 & 0.67 & 0.48 \\
Spotlighting                  & 0.68 & n/a & 0.39 & 0.77 & 0.00 & 0.51 & 0.31 \\
Adversarial training          & 0.41 & n/a & 0.29 & 0.18 & 0.04$^\dagger$ & n/a & 0.48 \\
Provenance monitor            & 0.41 & 0.00 & 0.29 & 0.00 & 0.54 & 0.67 & 0.48 \\
Mahalanobis OOD               & 0.41 & 0.02$^\dagger$ & 0.29 & 0.18 & 0.00$^\dagger$ & n/a & 0.48 \\
Width-based conformal         & 0.41 & 0.05$^\dagger$ & 0.29 & 0.18 & 0.24$^\dagger$ & n/a & 0.48 \\
\midrule
ConfGuard, $\Gconf$ only      & 0.38 & 0.08 & 0.26 & 0.16 & 0.08 & 0.09 & 0.45 \\
ConfGuard, $\Vphys$ only      & 0.00 & 0.00 & 0.00 & 0.18 & 0.88 & 0.88 & 1.00 \\
ConfGuard, both               & 0.00 & 0.08 & 0.00 & 0.16 & 0.12 & 0.12 & 0.92 \\
ConfGuard, both, no retry     & 0.00 & 0.46 & 0.00 & 0.11 & 0.08 & 0.09 & 0.45 \\
\B{ConfGuard $+$ provenance}  & 0.00 & 0.08 & 0.00 & 0.00 & 0.12 & 0.12 & 0.92 \\
\bottomrule
\end{tabular}
\end{table*}

\subsection{Complementary coverage holds, with unequal arms}
\begin{figure}[t]
\centering
\includegraphics{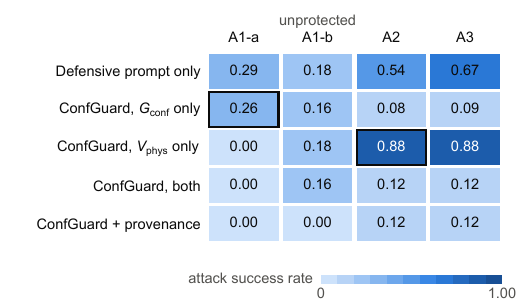}
\caption{Attack success rate by configuration on C-MAPSS FD001 with
a Qwen3-14B planner, the cells of Table~\ref{tab:main}. Outlined
cells show a removed gate's matched class returning: without
$\Vphys$, A1-a rises to 0.26; without $\Gconf$, A2 rises to 0.88.
A1-b lies outside the protected projection, and only composing a
provenance monitor removes it.}
\label{fig:ablation}
\end{figure}

Table~\ref{tab:main} reports FD001, the single-condition case that
isolates the mechanism, and reproduces with a real planner the
pattern Proposition~\ref{prop:complementary} predicts. Removing
$\Vphys$ returns A1-a to 0.26, close to the defended prompt's 0.29,
while A2 stays at 0.08. Removing $\Gconf$ sends A2 to 0.88 while
A1-a stays at 0.00. Each gate moves only its own class, and
Figure~\ref{fig:ablation} outlines the two cells that show it.

The $\Vphys$-only row is worse on A2 than no validator at all, 0.88
against the defended prompt's 0.54, and the cause is the retry.
Under A2 the corrupted interval warrants \textsc{maintain\_now} on
every attacked query. Left alone, the planner hedged to an
inspection on 102 of the 300, which happens to be safe; the
validator rejects that hedge as a departure from $\piref$, and the
retry hands the planner the class the corrupted interval warrants.
$\Vphys$ enforces the reference policy on whatever interval it is
given, as Proposition~\ref{prop:complementary}(b) states, which is
why only $\Gconf$ is credited with A2. Composed, the
system sits at 0.12 against 0.08 for $\Gconf$ alone, because the
attacked queries $\Gconf$ admits now dispatch the warranted class
instead of an inspection.

The retry earns that back on clean traffic. Without it, Qwen3-14B's
adherence failure of 0.41 becomes the validator's escalation rate,
and the composed system escalates 0.46 of clean decisions with a
utility of 0.45 under A1-a. With one retry, all 951 rejected
answers across all attacks came back in the warranted class. The
validator then escalates nothing on clean traffic, so clean
escalation falls to the 0.08 that $\Gconf$ contributes and utility
under A1-a rises to 0.92. Table~\ref{tab:main} reports both rows so
an operator can choose: the retry removes 0.38 of clean escalation
and adds 0.04 to A2.

The table also shows that the two arms differ in strength. The
$\Vphys$ column is binary because rejection follows from the
definition of a successful in-projection attack. The $\Gconf$
column stops at 0.08, above zero, as
Proposition~\ref{prop:power} anticipates: perturbations on
high-slack assets stay inside the support, and this mechanism
cannot detect them at any threshold.

\subsection{What the A1 number measures}
\label{sec:a1-meaning}

Read at face value, the 0.00 in the A1-a column suggests more than
the mechanism supports.

Assumption~\ref{as:eff} defines a successful in-projection attack
as $\Pi(a)\neq\piref(I(x))$, and \eqref{eq:vphys} rejects on
exactly that condition, so
Proposition~\ref{prop:complementary}(a) is true by construction.
The 0.00 is therefore not the rate at which attacks defeat a
mechanism that might have held. It is the rate at which our
\emph{implementation} fails to realise the construction, and on 300
attacked queries it failed on none. The implementation could fail
in two ways: a projection failure, where $\Pi$ cannot parse an
urgency class from a malformed call and the executor admits it, or
leakage on the retry path, where the second attempt escapes the
restricted action set. The planner produced no malformed call in
8{,}400, and a retried call is validated again before dispatch, so
the second cannot occur.

CaMeL makes a claim of the same kind \citep{debenedetti2025camel},
and security by construction is worth more than a large empirical
reduction. The two still have to be reported differently. Reporting
0.29 to 0.00 as an empirical defense would misdescribe what
happened: the comparison is between a mechanism that can be
persuaded and one that cannot be asked, and the correct figure of
merit is the residual and its causes.

The A2 column needs no such caveat. Its 0.08 for $\Gconf$ alone and
0.12 composed are measurements against a mechanism that can fail
and does, governed by the slack inequality of
Proposition~\ref{prop:power}.

\subsection{Sensor-side baselines, and why a detector is not a bound}
\label{sec:sensor-baselines}

Table~\ref{tab:sensorbase} compares the clock against the two
defences a reviewer is most likely to propose, a Mahalanobis detector
and adversarial training, and against width-based conformal
abstention. All run on the same windows, and every gate is calibrated
at the same $\rho$.

\begin{table}[t]
\centering
\caption{Sensor-channel defences on C-MAPSS FD001, all windows of
the 100 test engines. The universal perturbation of
\citet{basak2021universal} shifts predictions by $65.3$ cycles on
average and is a false-alarm attack: ungated, it flips $0.702$ of
healthy windows. The adaptive column is the same attacker
re-optimised to stay under each defence's own threshold, on 1{,}000
healthy windows for the learned defences and on every healthy
window for the clock, whose one-shot covert attacker is exact.}
\label{tab:sensorbase}
\small
\begin{tabular}{lccc}
\toprule
Defence & clean escalation & fixed A2 & adaptive \\
\midrule
None                     & 0.000 & 0.702 & n/a \\
Adversarial training     & n/a & 0.037 & 0.485 \\
Width-based conformal    & 0.050 & 0.235 & 0.925 \\
Mahalanobis OOD          & \B{0.019} & \B{0.000} & 0.504 \\
\B{Physics clock ($k{=}30$)} & 0.027 & 0.006 & \B{0.046} \\
\bottomrule
\end{tabular}
\end{table}

The paper's argument rests on the last two rows, read together.
Against the fixed perturbation the Mahalanobis detector
is perfect: it abstains on every attacked window at a clean cost of
$0.019$, better than ours. Against an attacker who knows its
threshold and optimises within it, it fails on $0.504$ of healthy
windows, with a median undetected shift of $54.1$ cycles and a
feasible evasion on \emph{every} window we tried. Adversarial
training behaves the same way: it absorbs the transferred attack
($0.037$) and then yields $0.485$ to a white-box one. The clock is
slightly worse than Mahalanobis on the fixed attack: it admits
$0.057$ of attacked healthy windows, and $0.006$ of healthy windows
reach a false alarm. Against the adaptive attacker it is an order
of magnitude better.

A learned detector's threshold sits on a score with no physical
meaning, so an attacker is free to move inside it, and nothing in
its construction bounds how far. The clock's threshold sits on a
quantity the asset itself constrains, so
Proposition~\ref{prop:power} bounds how far evasion can go, and the
measured $0.046$ is that bound's value on this data. Width-based
conformal is the weakest of the four. It stops only part of the
fixed attack, abstaining on $0.618$ of attacked windows and leaving
a false-alarm rate of $0.235$, and an attacker who keeps the
ensemble spread under the width threshold reaches a false alarm on
$0.925$ of healthy windows. We do not claim the clock detects more
attacks than a well-tuned OOD detector, only that it is the one
defence of the four whose evasion margin can be stated in advance.

\subsection{The gates beat the obvious alternatives}
\label{sec:alternatives}
On the sensor channel the comparison is Table~\ref{tab:sensorbase}.
The width-based gate is the closest relative of $\Gconf$, so it
isolates what the support test in \eqref{eq:score} adds: on the same
windows it lets $0.235$ of healthy windows reach a false alarm under
the fixed attack against the clock's $0.006$, and $0.925$ under the
adaptive one against $0.046$. Adversarial training of the backbone
\citep{siddique2023robustpdm} costs accuracy, as it usually does
(test RMSE 14.1 cycles against 13.5). It defends the regressor,
not the decision, so it composes with a monitor instead of
competing with it.

Two rows of Table~\ref{tab:main} are easy to misread. The
prompt-only planner's $0.00$ on A2 is not robustness. Without the
decision rule, Qwen3-14B answered \texttt{schedule\_overhaul} on
all 300 clean queries, which the harm model never counts as
harmful, so an attack on the prognostic has nothing to move; the
same behaviour produces its clean decision error of $0.68$. A text
injection does move it, to $0.39$ on A1-a and
$0.52$ on A3. Spotlighting \citep{hines2024spotlighting} changes
none of this on this model: $0.39$ against $0.39$ on A1-a and $0.77$
against $0.76$ on A1-b.

The provenance monitor is the informative comparison. Run behind
the same rule prompt as ConfGuard, it leaves A2 at $0.54$, exactly
the defended prompt's figure, because a capability policy has no
predicate that can call a prognostic number physically implausible:
the tool call is well-formed, the tool is permitted, and the
argument's provenance is clean. This row measures the gap the paper
claims. The same monitor takes A1-b from $0.18$ to $0.00$ by
escalating every call that carried an injected part, technician or
service window, because corrupting a part number \emph{is} a
provenance violation and falls inside what a capability policy
sees. It does nothing for A1-a, which stays at $0.29$.

\subsection{The static support misses the attack that matters}
\label{sec:static-fails}

\begin{figure*}[t]
\centering
\includegraphics[width=\linewidth]{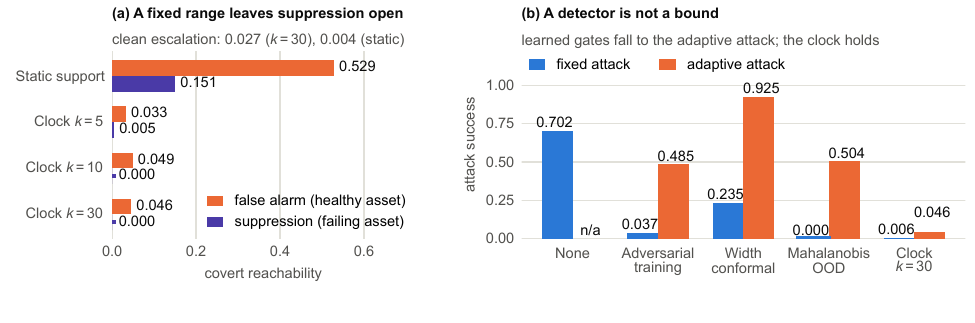}
\caption{C-MAPSS FD001 test engines, thresholds from
Proposition~\ref{prop:calib}. (a) Covert reach on all 10{,}196 test
windows (Table~\ref{tab:clock}). A fixed physical range is blind to
suppression and leaves a covert attacker reaching $0.529$ of healthy
windows; anchoring to the asset's own history cuts false-alarm reach
to $0.046$ at $k{=}30$, and at $k{=}10$ and $k{=}30$ closes
suppression. (b) Fixed and adaptive attacks (Table~\ref{tab:sensorbase}).
A Mahalanobis detector, adversarial training and the clock all stop
the fixed universal perturbation; the first two fall to an attacker
who re-optimises under their own threshold, measured on 1{,}000
healthy windows for the learned defences and on every healthy window
for the clock. No adaptive attack applies to the undefended
pipeline. Only the physics clock carries a bound into the adaptive
setting (Proposition~\ref{prop:power}).}
\label{fig:clock}
\end{figure*}

Table~\ref{tab:clock} is the measurement that set the design, and
Figure~\ref{fig:clock} sets it beside the sensor baselines of
Section~\ref{sec:sensor-baselines}. It reports, per monitor, the
fraction of windows at which a covert attacker (one that stays
inside the admitted interval of Section~\ref{sec:gconf}) can still
reach a harmful decision class.
Because every score is monotone in the prediction, that fraction is
computed exactly rather than by search: the attacker's best move is
the end of the admitted interval.

\begin{table}[t]
\centering
\caption{Covert reachability on all 10{,}196 windows of the 100
C-MAPSS FD001 test engines. Each threshold is the per-window shape
$\tau^0$ from the ten held-out validation engines, widened by the
$\hat\lambda$ of Proposition~\ref{prop:calib}, cross-fit over two
halves of the test engines (last column, one value per half).
$\hat q = 27.89$ cycles, $\rho = 0.05$, attacker budget $65$ cycles.
Suppression is reachable when the attacker can raise a critical
asset ($y<25$) into \textsc{no\_action}; false alarm when it can
lower a healthy asset ($y>60$) into \textsc{maintain\_now}.}
\label{tab:clock}
\small
\begin{tabular}{lcccc}
\toprule
Monitor & clean escalation & suppression & false alarm & $\hat\lambda$ \\
\midrule
Static support        & \B{0.004} & 0.151 & 0.529 & 1.02, 1.02 \\
Clock, $k{=}5$        & 0.034 & 0.005 & \B{0.033} & 1.44, 1.40 \\
Clock, $k{=}10$       & 0.030 & \B{0.000} & 0.049 & 1.18, 1.30 \\
\B{Clock, $k{=}30$}   & 0.027 & \B{0.000} & 0.046 & 1.00, 1.00 \\
\bottomrule
\end{tabular}
\end{table}

Calibrated to the same $\rho$, the static support escalates almost
no clean windows ($0.004$) and leaves a covert attacker reaching
$0.529$ of healthy windows on false alarm and $0.151$ of critical
ones on suppression. The $k{=}30$ clock spends $0.027$ of clean
windows, inside the $0.05$ target, to cut false-alarm reach to
$0.046$, a factor of $11.5$, and to close suppression. Once each
horizon is calibrated over trajectories the three sit close
together: $k{=}5$ reaches $0.033$ on false alarm at $0.034$
escalation but leaves $0.005$ of suppression open, and $k{=}10$
reaches $0.049$ at $0.030$. We use $k{=}30$ for the rest of the
paper because it closes suppression at the lowest clean cost of the
three, and because a short history is easier for a persistent
attacker to shift (Section~\ref{sec:catch}).

Calibrating over trajectories changed the first three rows. A
per-window calibration, which ranks window scores as if they were
exchangeable, puts $k{=}5$ at $0.140$ of clean test windows and the
static support at $0.058$, both above the $\rho$ they were
calibrated to. Proposition~\ref{prop:calib} widens those thresholds
until the
trajectory-level bound holds and leaves $k{=}30$ where it was
(Section~\ref{sec:falseabst}). On the ten validation engines,
cross-fit so thresholds never score the engines that set them, the
per-window comparison gives $0.626$ against $0.082$ at matched
escalation ($0.045$ and $0.046$), so the ordering does not depend on
the split. Ten engines are too few for Proposition~\ref{prop:calib}
at $\rho=0.05$, which needs 19.

The static row's suppression figure is bounded by the attacker's
$65$-cycle budget, not by the support, which is blind in that
direction at any budget. Nor does the clock gain from a looser
floor: at $k{=}30$ the floor threshold $\tau_F$ is $24.02$ cycles,
below the static support's $24.50$, so the entire difference comes
from the anchors in \eqref{eq:anchors}.

\subsection{False abstention under the sequential monitor}
\label{sec:falseabst}

Proposition~\ref{prop:calib} bounds the expected share of a clean
asset's windows that the deployed monitor escalates.
Table~\ref{tab:falseabst} measures that share on the FD001 and FD004
test engines, before and after the calibration over trajectories.
The per-window calibration misses its target. On FD001 it
escalates $0.140$ of clean windows at $k{=}5$, $0.077$ at $k{=}10$
and $0.058$ under the static support, and at $k{=}5$ the asset at
the 90th percentile loses a third of its windows. The cause is the
one Section~\ref{sec:gconf} names: a rejected window stays out of
the history, so one rejection can start a run of them. After the
calibration over trajectories every monitor sits under $\rho$ on
both datasets, per asset and pooled. At $k{=}30$ the per-window
thresholds already met the bound, so both halves choose
$\hat\lambda=1$ on both datasets and nothing reported elsewhere for
$k{=}30$ changes. MetroPT-3 is outside the proposition, with two
trajectories where 19 are needed, and its per-window figure of
$0.075$ is reported as measured.

\begin{table*}[t]
\centering
\caption{Clean false abstention of the deployed monitor (a window
enters the history only if it passes) on every test window, at
$\rho=0.05$. \emph{Per-window}: thresholds from per-window quantiles
on the validation engines. \emph{Proposition~\ref{prop:calib}}: the
same thresholds widened by $\hat\lambda$, chosen on one half of the
test engines and applied to the other (one value per half).
\emph{Pooled} is the share of all clean windows escalated;
\emph{mean} and \emph{p90} are taken over assets, each asset's own
share, so \emph{mean} is the quantity the proposition bounds.}
\label{tab:falseabst}
\small
\begin{tabular}{ll ccc cccc}
\toprule
& & \multicolumn{3}{c}{Per-window} & \multicolumn{4}{c}{Proposition~\ref{prop:calib}}\\
\cmidrule(lr){3-5}\cmidrule(lr){6-9}
Data & Monitor & pooled & mean & p90 & $\hat\lambda$ & pooled & mean & p90 \\
\midrule
FD001, 100 engines & static support & 0.058 & 0.058 & 0.124 & 1.02, 1.02 & 0.004 & 0.004 & 0.000 \\
                   & clock, $k{=}5$  & 0.140 & 0.115 & 0.334 & 1.44, 1.40 & 0.034 & 0.030 & 0.084 \\
                   & clock, $k{=}10$ & 0.077 & 0.065 & 0.194 & 1.18, 1.30 & 0.030 & 0.027 & 0.092 \\
                   & clock, $k{=}30$ & 0.027 & 0.023 & 0.076 & 1.00, 1.00 & 0.027 & 0.023 & 0.076 \\
\midrule
FD004, 248 engines & static support & 0.061 & 0.059 & 0.196 & 1.02, 1.02 & 0.000 & 0.000 & 0.000 \\
                   & clock, $k{=}10$ & 0.066 & 0.066 & 0.230 & 1.30, 1.02 & 0.045 & 0.044 & 0.137 \\
                   & clock, $k{=}30$ & 0.030 & 0.029 & 0.061 & 1.00, 1.00 & 0.030 & 0.029 & 0.061 \\
\midrule
MetroPT-3, 2 episodes & clock, $k{=}10$ & 0.075 & n/a & n/a & \multicolumn{4}{c}{not applicable: needs 19 trajectories} \\
\bottomrule
\end{tabular}
\end{table*}

The real fleets cannot be varied on demand, so Table~\ref{tab:sim}
repeats the test on simulated ones. Each scenario draws whole
trajectories: true capped remaining life over a lifetime of 128 to
362 cycles, and a prediction whose error follows a named process.
\emph{Smooth} is an AR(1) error with standard deviation 13 cycles
and lag-one correlation 0.95, close to the FD001 regressor.
\emph{Noisy} has standard deviation 20 and correlation 0.7.
\emph{Abrupt onset} holds the prediction at the cap until true life
falls below a random level between 30 and 90 cycles, then drops to
track it. \emph{Switching regimes} adds one of six operating-condition
offsets, from $-12$ to $+18$ cycles, that switches with probability
0.05 per cycle. \emph{Rate mismatch} lets each asset degrade up to
30\% faster or slower than the predictor learned. Trajectories are
cut at a random point, as in the C-MAPSS test set, or observed to
failure. Each repetition sets $\tau^0$ by per-window calibration on
10 run-to-failure trajectories, as on FD001, chooses $\hat\lambda$ on
50 trajectories of the deployment kind, and runs the real monitor at
$k{=}30$ on 400 fresh ones; the table averages 10 repetitions.

\begin{table*}[t]
\centering
\caption{Simulated fleets, monitor at $k{=}30$, $\rho=0.05$, 10
repetitions of 400 test trajectories each. \emph{Mean} is the
per-asset false-abstention share averaged over assets and
repetitions, and \emph{worst} the largest per-repetition mean.
Covert reach is the exact one-shot false-alarm reach on healthy
windows (budget 65 cycles), the detection price of widening the
thresholds. The last row calibrates on one pattern and deploys on
another, which breaks the assumption of
Proposition~\ref{prop:calib}.}
\label{tab:sim}
\small
\begin{tabular}{ll cc ccc cc}
\toprule
& & \multicolumn{2}{c}{Per-window} & \multicolumn{3}{c}{Proposition~\ref{prop:calib}} & \multicolumn{2}{c}{Covert reach}\\
\cmidrule(lr){3-4}\cmidrule(lr){5-7}\cmidrule(lr){8-9}
Pattern & Observed & mean & worst & $\hat\lambda$ & mean & worst & per-window & Prop.~\ref{prop:calib} \\
\midrule
Smooth            & truncated & 0.046 & 0.072 & 1.09 & 0.034 & 0.055 & 0.049 & 0.055 \\
Noisy             & truncated & 0.031 & 0.040 & 1.03 & 0.027 & 0.036 & 0.266 & 0.285 \\
Abrupt onset      & truncated & 0.023 & 0.045 & 1.04 & 0.021 & 0.045 & 0.631 & 0.659 \\
Switching regimes & truncated & 0.038 & 0.074 & 1.08 & 0.027 & 0.046 & 0.049 & 0.055 \\
Rate mismatch     & truncated & 0.030 & 0.041 & 1.03 & 0.026 & 0.041 & 0.036 & 0.038 \\
Smooth            & to failure & 0.098 & 0.137 & 1.40 & 0.033 & 0.046 & 0.076 & 0.123 \\
\midrule
Smooth $\to$ regimes & truncated & 0.059 & 0.097 & 1.12 & 0.040 & 0.061 & 0.035 & 0.043 \\
\bottomrule
\end{tabular}
\end{table*}

Whenever the calibration and deployment fleets are drawn the same
way, the calibrated monitor escalates between 0.021 and 0.034 of a
clean asset's windows on average, under $\rho=0.05$ for every
pattern. The largest single repetition reaches 0.055, which a bound
on the expectation allows. The per-window calibration averages
under $\rho$ on truncated fleets, but its worst repetitions reach
0.072 and 0.074, and on run-to-failure fleets, where every
trajectory crosses end of life, it averages 0.098. The proposition
brings that to 0.033 by widening the thresholds ($\hat\lambda=1.40$
on average), and the widening has a detection price: one-shot covert
reach rises from 0.076 to 0.123. On the noisy and abrupt patterns
covert reach is high under either calibration, because a poor
predictor widens $\hat q$ and with it every admitted interval; that
is the predictor's cost, not the calibration's. The last row breaks
the assumption on purpose. Calibrated on the smooth pattern and
deployed on switching regimes, the monitor averages 0.040, but one
repetition reaches 0.061, and nothing in the proposition covers it.
The calibration fleet has to come from the regimes the monitor will
see.

\subsection{What the monitor catches before harm}
\label{sec:catch}

Table~\ref{tab:catch} measures whether, once an attack arrives, the
monitor escalates before the first harmful dispatch. It is
evaluated along full trajectories, with the
attacker persistent from the degradation onset, so the monitor's
history is the one the attacker shaped.

\begin{table}[t]
\centering
\caption{Sensor attacks along full trajectories on all 100 FD001
test engines, attacker persistent from the degradation onset.
\emph{Caught} is the fraction of attacked assets escalated before
any harmful dispatch; \emph{FA} is the false-alarm success rate
over healthy windows. The covert attacker is idealised: it may
place the prediction anywhere in the $\epsilon$-ball's image and
re-plans every window, an upper bound on attack power.}
\label{tab:catch}
\small
\begin{tabular}{llcc}
\toprule
Attack & Monitor & caught before harm & FA success \\
\midrule
\multirow{2}{*}{Universal perturbation}
 & static support & 0.000 & 1.000 \\
 & clock, $k{=}30$ & \B{1.000} & \B{0.142} \\
\midrule
\multirow{2}{*}{Adaptive, non-covert}
 & static support & 0.000 & 1.000 \\
 & clock, $k{=}30$ & \B{1.000} & \B{0.138} \\
\midrule
\multirow{2}{*}{Adaptive, covert, persistent}
 & static support & 0.000 & 1.000 \\
 & clock, $k{=}30$ & 0.000 & 0.638 \\
\bottomrule
\end{tabular}
\end{table}

In the first two blocks the defence works as designed. The
universal perturbation and the non-covert adaptive attacker are
caught before harm on every asset under the clock, and on none
under the static support, which never sees a downward shift that
stays inside $[0,125]$. False-alarm success falls from
$1.000$ to $0.142$ on the universal attack.

The third block shows the limit. A covert attacker who re-plans
every window and holds each step inside the admitted interval
drifts through: $0.638$ of healthy windows reach a false alarm
under the $k{=}30$ clock against $1.000$ under the static support.
The mechanism is history poisoning. Each admitted step enters
$H_t$, so the anchors follow the attacker down at up to $\tau_T$
per window. Shorter horizons do worse here, because a short history
is easier to shift: $0.868$ at $k{=}5$ and $0.848$ at
$k{=}10$.

Against this attacker the clock still guarantees a rate, which the
alternatives do not. Proposition~\ref{prop:power}
bounds each admitted step by $m(t)+\tau_\rho$, and with the clock
$m_{\downarrow}$ is at most $\tau_T$ once the history is the
attacker's, so a healthy asset predicted at the cap needs at least
$(125-\theta_{\mathrm{d}}-\hat q)/(\tau_T+1)\approx 3.4$ windows to
reach a false alarm at $k{=}30$, and the drift leaves a visible
monotone trace in the admitted history. Section~\ref{sec:limitations}
names the obvious repair, anchoring to the last human-verified
state rather than to the rolling admitted window.

Suppression has a simpler explanation. The covert suppression rate
is $0.221$ under every monitor, including none, because that is as
far as the $\epsilon$-ball reaches: the median upward movement PGD
can induce on a test window is $8.8$ cycles against $67.0$
downward. The regressor is far easier to push toward failure than
away from it, so on this backbone suppression is bounded by the
attack, not by the gate.

\subsection{Across datasets and planners}

Table~\ref{tab:perdataset} splits the headline cells by dataset. FD004, with six operating conditions and two fault modes, is the case Mondrian conformal prediction was designed for. On all 248 FD004 test engines, with thresholds from Proposition~\ref{prop:calib}, the static support leaves a covert attacker suppression on 0.808 of critical windows and false alarm on 0.221 of healthy ones; the $k{=}30$ clock cuts these to 0.099 and 0.038 while escalating 0.030 of clean windows. Contrary to our expectation, switching the conformal width from pooled to per-condition Mondrian makes no difference: false alarm moves to 0.039 and suppression to 0.098. The clock's reference is the asset's own admitted history, and a regime shift moves the interval width but not the trajectory, so the monitor tolerates changing operating conditions without a per-regime calibration. Mondrian still matters for the floor term and for the interval the planner reads, but the clock does not depend on it. Suppression stays further from zero than on FD001 because six regimes make the predictor noisier along a trajectory, and the anchors carry that noise as a wider band; a shorter horizon does not close it (0.085 at $k{=}10$, at 0.045 clean escalation). The universal perturbation is the weak point of the FD004 evidence, and the attack is the cause, not the gate: fitted against this model, it moves test RMSE only from 14.9 to 19.5 and flips 0.049 of healthy windows. Against it the clock catches 0.539 of 154 attacked assets before harm, against 0.006 for the static support, so the covert-reachability figures in Table~\ref{tab:perdataset} are the fairer comparison across datasets.

MetroPT-3 is a different regime, and the clock behaves differently on it. Its target is remaining life in days capped at 14, sampled every ten minutes, so remaining life moves by 0.007 per window and a fixed range is useless: a covert attacker reaches a false alarm on 0.998 of healthy windows under the static support. The clock holds it to 0.104 with a ten-window history and 0.202 with thirty, the reverse of the C-MAPSS ordering, because thirty ten-minute windows is a five-hour anchor on a signal that has barely moved. Suppression is only partly closed, 0.500 to 0.183 over 101 critical windows: the calibrated tolerance $\tau_T$ is 2.5 days on this predictor, wider than the 2-day maintenance threshold it protects. The horizon $k$ is a quantity of time, not of windows, and a deployment should set it from the asset's degradation rate. Proposition~\ref{prop:calib} does not apply here. The test stream is one unit with two failure episodes, two trajectories where $\rho=0.05$ needs 19, so MetroPT-3 keeps per-window thresholds cross-fit across its two episodes, and clean abstention sits at 0.075, above the 0.05 target. A deployment would calibrate on a fleet of such units, or on many more episodes of this one.

\begin{table}[t]
\centering
\caption{One-shot covert reach by dataset, static support against the
physics clock, at the clean escalation rate each clock pays. All cells are
sensor-side and need no planner; the planner sweep is FD001 only
(Table~\ref{tab:main}). C-MAPSS rows score every test window, with
thresholds from Proposition~\ref{prop:calib} cross-fit over halves of
the test engines. MetroPT-3 has two episodes, too few for the
proposition, and keeps per-window thresholds cross-fit across them.
The horizon is $k{=}30$ windows on C-MAPSS and $k{=}10$ on MetroPT-3,
whose ten-minute windows make thirty a five-hour anchor.}
\label{tab:perdataset}
\small
\begin{tabular}{lccccc}
\toprule
& & \multicolumn{2}{c}{Suppression $\downarrow$} & \multicolumn{2}{c}{False alarm $\downarrow$}\\
\cmidrule(lr){3-4}\cmidrule(lr){5-6}
Dataset & clean abst & static & clock & static & clock \\
\midrule
C-MAPSS FD001 & 0.027 & 0.151 & \B{0.000} & 0.529 & \B{0.046} \\
C-MAPSS FD004 & 0.030 & 0.808 & \B{0.099} & 0.221 & \B{0.038} \\
MetroPT-3 & 0.075 & 0.500 & \B{0.183} & 0.998 & \B{0.104} \\
\bottomrule
\end{tabular}
\end{table}

Assumption~\ref{as:adh} is the only empirical assumption in the
analysis, and $\rho_\pi$ is a property of the planner rather than of
ConfGuard, so one model family would not settle it.
Table~\ref{tab:planners} measures it on five open planners from five
model families, and none follows a three-line threshold rule
reliably: first-answer adherence failure on clean queries runs from
0.41 (Qwen3-14B) to 0.65 (Mistral-Nemo-12B). The failures are
systematic: between 82\% and 100\% of each planner's deviations are
hedges to an inspection where the rule asks for an extreme class,
and the worst-class rate reaches 1.00 for Phi-4-14B and
Mistral-Nemo-12B, which never issue \textsc{no\_action} when the rule
requires it.

This has two consequences. First, the clean average does not bound
the validator's trigger rate under A2, because A2 moves every query into
one warranted class: for Gemma-3-12B and Mistral-Nemo-12B the
validator triggers on 0.82 and 0.70 of attacked queries against
clean averages of 0.63 and 0.65. The worst-class rate bounds it on
all five, which is why Assumption~\ref{as:adh} is stated per class.
Second, the retry absorbs the failure: after one corrective call the
validator escalates no clean query for any of the five planners,
and the A1-a residual is 0.00 on all five. Neither the labour cost
nor the text-channel guarantee depends on which planner sits behind
the monitor. The undefended exposure does: with the rule prompt
alone, A1-a runs from 0.01 for Phi-4-14B to 0.31 for Llama-3.1-8B,
and the validator removes that spread.

\begin{table}[t]
\centering
\caption{Planner adherence on FD001 under the rule prompt, five open
planners, three seeds of 100 queries per attack. $\rho_\pi$ is the
first-answer adherence failure on clean queries and \emph{worst} the
same rate in the warranted class where it is highest; the next column
is $\Vphys$'s trigger rate under A2, which
Proposition~\ref{prop:complementary}(b) bounds by the worst-class
rate. \emph{After retry} is the clean escalation rate of $\Vphys$ once
the bounded retry has run.}
\label{tab:planners}
\small
\setlength{\tabcolsep}{4pt}
\begin{tabular}{lcccccc}
\toprule
& \multicolumn{2}{c}{$\rho_\pi$, clean} & $\Vphys$ trig. & Clean abst & \multicolumn{2}{c}{A1-a ASR}\\
\cmidrule(lr){2-3}\cmidrule(lr){6-7}
Planner & all & worst & under A2 & after retry & prompt & $+\Vphys$ \\
\midrule
Qwen3-14B & 0.410 & 0.723 & 0.340 & 0.000 & 0.287 & 0.000 \\
Phi-4-14B & 0.500 & 1.000 & 0.393 & 0.000 & 0.013 & 0.000 \\
Gemma-3-12B & 0.627 & 0.864 & 0.820 & 0.000 & 0.117 & 0.000 \\
Mistral-Nemo-12B & 0.648 & 1.000 & 0.703 & 0.000 & 0.090 & 0.000 \\
Llama-3.1-8B & 0.456 & 0.907 & 0.360 & 0.000 & 0.310 & 0.000 \\
\bottomrule
\end{tabular}
\end{table}

\subsection{Adaptive attacks}
The adaptive attacker of Section~\ref{sec:adaptive} knows every
threshold. Two of its forms are already in Table~\ref{tab:catch}.
The non-covert form, which maximises damage and ignores the
monitor, is caught before harm on every asset. The covert form
that re-plans every window drifts through, at $0.638$; the
one-shot covert attacker, who gets a single
window against a clean history, is bounded exactly by
Proposition~\ref{prop:power}, and the self-check that the
admitted interval and the pass decision agree held on $1{,}000$
random monitor states with no failures.

Against A4-t the attack succeeds on 0.00 of queries, if success
means changing the dispatched decision: every injection that passes
$\Vphys$ requests an action already endorsed by $\piref(I(x))$, so
the dispatched decision is the one the operator would have taken
without the attack. That holds by construction, so there is no
attack rate to measure. The A1-a residual of
Section~\ref{sec:a1-meaning}, 0.00 on 300 attacked queries with no
malformed call in 8{,}400, measures instead whether the
implementation realises the construction. A policy-consistent
injection can still corrupt the
fields outside the projection, which is A1-b.

\subsection{Forced abstention}
\label{sec:dos}

Proposition~\ref{prop:calib} bounds the expected abstention rate on
\emph{clean} trajectories exchangeable with the calibration fleet.
It says nothing about the rate under attack, and an attacker can
exploit that asymmetry. An
attacker with sensor-channel access who wants disruption rather
than a decision flip can invert the A4-s objective: instead of
maximising prediction shift subject to staying under $\tau_\rho$,
drive $s(x+\delta)$ above $\tau_\rho$ on every window. Nothing in
our construction prevents this, and we measure the result: a
perturbation fitted to maximise the support-exit score forces
abstention on 0.961 of queries at $k{=}30$ (and 0.988 at $k{=}5$), so the agent escalates almost
everything and the maintenance planners absorb the load.

The failure is one of availability, not decision integrity: the
agent dispatches nothing wrong, and in that narrow sense the
monitor works as specified. It follows directly from choosing
escalation as the response to detection. A monitor that blocked
silently would not have this exposure, but it would not be
deployable either, because an industrial planner needs somewhere
for an uncertain decision to go.

We do not defend against this in the present work, and the
escalation path is not free. Two observations bound how bad it is.
First, forced abstention is loud: an attacker pursuing it gives up
the covertness that makes the flip attacks dangerous, and the
operator learns the agent is under attack rather than being quietly
misled. Second, and more usefully, the calibration that creates the
exposure also gives a reference for detecting it.
Under Proposition~\ref{prop:calib}, clean assets escalate at most
$\rho$ of their windows on average, and the calibration fleet shows
how that rate spreads across assets (Table~\ref{tab:falseabst}), so
a fleet-wide rate well above that spread is evidence of attack.
Turning that observation into a sequential test with a controlled
false-alarm rate, and pairing it with per-asset escalation rate
limits, is the obvious next step, and we have not evaluated it.

\subsection{The class we do not defend}
A1-b sits at 0.16 for full ConfGuard against 0.18 for the defended
prompt it runs on. The defense has almost no effect, as
Section~\ref{sec:why-planner} predicts, because those injections
never touch the projected coordinate; the drop that does appear
comes from $\Gconf$ escalating 0.08 of queries for unrelated
reasons. Composing the provenance monitor takes it to 0.00 at an
escalation rate of 0.24 on those queries. An operator deploying
ConfGuard is protected on maintenance urgency and unprotected on
part selection and scheduling, and should compose a provenance
monitor such as Progent \citep{shi2025progent} to cover the rest.

\subsection{Cost of the defense}
On clean traffic the composed system escalates 0.08 of decisions,
all of it from $\Gconf$, since the retry leaves the validator with
none. That is above the $\rho=0.05$ set at calibration because the
sweep draws its queries from windows at or after the degradation
onset, where the predictor is noisier than on the calibration mix;
on all test windows the same clock escalates 0.027
(Table~\ref{tab:clock}). Utility under A1-a rises from 0.48 for the
defended prompt, and 0.32 undefended, to 0.92, because a rejected
hijack is corrected on the retry rather than dispatched.

The bill follows Section~\ref{sec:exchange}. Clean cost per
decision is \$6.90 for the composed system: \$3.60 of escalation
and the rest the reference policy's own false alarms, which are
preventive maintenance it orders on intervals that cross
$\theta_{\mathrm{d}}$. Under the hybrid attack A3 the composed system
costs \$55 per decision against \$587 for the defended prompt and
\$1{,}065 undefended. Under A2 alone the undefended planner looks
free, \$0, only because it never acts on the prognostic and the cost
model does not charge its inspections. The retry is expensive under
covert suppression. The attacker raises the prediction of a
failing asset inside the admitted interval; without the retry the
planner hedges and the validator escalates, at \$205 per decision,
and with it the planner is handed \textsc{no\_action}, at \$840 per
decision against \$174 for $\Gconf$ alone. The rate is small, 0.025
of attacked queries, but each success is a \$50k failure. Dropping the
retry only for assets the clock places near $\theta_{\mathrm{d}}$
would keep most of the clean saving, and we have not evaluated it.
Under the forced-abstention attack of Section~\ref{sec:dos} the sign
flips again, to \$43 per decision in pure escalation labour with no
maintenance value delivered, which is the number an operator should
see before deploying this.

The checks add no latency an operator would notice. On one Kaggle
CPU core the $k{=}30$ clock takes 69\,$\mu$s per window, the
validator 17\,$\mu$s to parse, project and compare a real planner
answer, and the provenance check 19\,$\mu$s, about
105\,$\mu$s per decision in all. The planner call it guards
has a median latency of 22.8\,s (90th percentile 26.6\,s) on a T4
serving four requests at once, so the monitor adds about
$5\times10^{-6}$ of the loop. A $\Gconf$ escalation also skips the
planner call entirely.

\section{Limitations}
\label{sec:limitations}

\paragraph{Coverage is bounded by the projection} $\Vphys$ protects
one three-valued coordinate. A1-b measures the consequence: 0.16
for ConfGuard alone (Table~\ref{tab:main}).
Extending the projection to part numbers and scheduling needs a
reference policy over those fields, which is inventory logic rather
than physics, and we have not written it.

\paragraph{The support test needs a real support} $\Gconf$ requires
a physically meaningful $S$. RUL has a hard floor at zero, which is
what makes the C-MAPSS case clean. Quantities without a natural
bound get a weaker version of the same test, and the slack argument
degrades accordingly.

\paragraph{Interval-valued prognostics only} The mechanism does not
transfer to classification. A binary failure predictor such as AI4I
2020 \citep{matzka2020ai4i} produces a probability, which has a
support of $[0,1]$ that every prediction satisfies by definition, so
\eqref{eq:score} is identically zero and $\Gconf$ carries no
signal. Substituting a probability threshold would be a different
mechanism under the same name, so we excluded the dataset rather
than report it. Extending the construction to classification needs
a conformal object with a physically meaningful exterior, for
instance a predicted time-to-event rather than a class posterior,
and we have not built it.

\paragraph{Censored targets weaken the guarantee} MetroPT-3 shows
the boundary quantitatively: a target capped at 14 days leaves
slack nearly constant across assets, and
Corollary~\ref{cor:risk} needs slack to vary with asset health to
deliver risk-weighted detection.

\paragraph{The rolling history can be poisoned} Section~\ref{sec:catch}
measures it: a covert attacker who re-plans every window and keeps
each step inside the admitted interval reaches a false alarm on
$0.638$ of healthy windows, because admitted values are the
anchor. The clock turns an unbounded one-shot attack into a bounded
drift, which is a real guarantee, though not the one an operator
most wants. The repair is to anchor $A^{\uparrow},A^{\downarrow}$ to a
reference the attacker cannot write, the last human-verified
inspection or the trajectory before onset, and to treat a monotone
run of admitted steps as its own signal. We evaluated that repair in the same run. Committing the anchors only at an inspection event every 25 cycles lowers the persistent covert attacker's reach from 0.638 to 0.480, at a clean escalation rate of 0.027 that is itself lower than the rolling anchor's, because the attacker can no longer move the reference between inspections. Every 50 cycles gives 0.655, worse than rolling: a snapshot that old has stopped tracking the asset, so the interval of admissible predictions widens again. The committed anchor is a partial repair with a tunable inspection cadence, and the full one remains a reference the attacker cannot write.

\paragraph{The calibration guarantee needs a calibration fleet}
Proposition~\ref{prop:calib} holds in expectation over assets that
are exchangeable with the calibration fleet as whole trajectories,
and it needs at least $1/\rho-1$ of them. It says nothing when the
fleet's regime mix changes after calibration
(Table~\ref{tab:sim}, last row) or when one unit is all there is,
as on MetroPT-3. Recalibrating $\hat\lambda$ as the fleet changes is
the operational remedy; a bound under drift needs a weighted form of
the argument \citep{barber2023beyond} that we have not derived.

\paragraph{Availability is not defended} Section~\ref{sec:dos}
measures a forced-abstention attack that stops the agent working at
0.961 of queries. We give a detection argument for it and no
defense. Any deployment needs an escalation rate limit and a
policy for what the loop does when the monitor is saturated, and
neither is evaluated here.

\paragraph{Static conformal calibration} We recalibrate offline. An
asset whose degradation regime drifts will see $\hat q$ go stale,
inflating slack and weakening detection. Adaptive conformal
inference would address this and we have not evaluated it.

\paragraph{Open planners only} The main results use one open-weight
14B planner, and the adherence measurement adds four more, all
between 8B and 14B parameters. We ran no hosted frontier model. The
confinement argument for A4-t does not depend on the planner, but
the adherence rate $\rho_\pi$ in Assumption~\ref{as:adh} does, and
with it the clean escalation cost of $\Vphys$
(Table~\ref{tab:planners}).

\paragraph{Synthetic injection corpus} The adversarial chunks are
generated from templates, not collected from real compromised
documentation. The encodings are drawn from published attacks, but
the distribution is ours.

\section{Conclusion}
\label{sec:conclusion}

Agent reference monitors currently check where data came from. In
an industrial maintenance loop the security-critical quantity is
physical, and a policy grounded in the physics of the asset gives
guarantees a provenance policy cannot. Testing each prediction
against the asset's own history under its degradation physics, with
a threshold calibrated over whole clean trajectories, gives a monitor
whose expected clean escalation rate the operator sets directly for
the monitor as it runs, and whose detection power rises as the asset
approaches failure. Comparing the agent's urgency decision against a
deterministic policy over the same interval makes in-projection
injection unreachable by construction and confines an adaptive text
attacker to actions the physics policy already endorses; what
remains is an implementation residual, which we report as such.

Neither mechanism eliminates its channel. The sensor monitor bounds
the attacker by a slack the operator can read off the prediction,
and the validator bounds the text attacker to decisions that were
already correct. Both leave classes uncovered, and the provenance
monitor covers several of them, which argues for composing the two
families rather than choosing between them. Neither addresses
availability. Answering a detected attack by calling a human makes
the monitor deployable and also makes forced escalation worth
attempting; we price that channel instead of excluding it. Now that
adaptive evaluation has become the standard
\citep{zhan2025adaptive}, a defense has to be stated as the bound it
achieves against an attacker who knows the thresholds.

\section*{Declaration of competing interest}
The authors declare that they have no known competing financial
interests or personal relationships that could have appeared to
influence the work reported in this paper.

\section*{Data availability}
Code, configurations, trained checkpoints and result files are
available from the corresponding author on request.
The datasets are public: C-MAPSS FD001 and FD004
\citep{saxena2008cmapss} and MetroPT-3
\citep{veloso2022metropt}.

\bibliographystyle{elsarticle-num-names}
\bibliography{refs}

\end{document}